\documentclass{article}

\usepackage{PRIMEarxiv}

\usepackage[french,german,spanish,russian,main=english]{babel}
\usepackage[T1]{fontenc}    % use 8-bit T1 fonts
\usepackage{hyperref}       % hyperlinks
\usepackage{url}            % simple URL typesetting
\usepackage{booktabs}       % professional-quality tables
\usepackage{amsfonts}       % blackboard math symbols
\usepackage{nicefrac}       % compact symbols for 1/2, etc.
\usepackage{microtype}      % microtypography
\usepackage{lipsum}
\usepackage{fancyhdr}       % header
\usepackage{graphicx}       % graphics
\usepackage{subcaption}
\usepackage{footnote}
\usepackage{tabularx}
\makesavenoteenv{table}
\makesavenoteenv{tabular}

\usepackage{array,  makecell}
\usepackage{longtable}
\usepackage[labelformat=simple]{subcaption}	

\usepackage{multirow}
\usepackage{utfsym}
\usepackage{algorithm}
\usepackage{algpseudocode}
\usepackage{mathtools}
\usepackage{cmap}
\usepackage{pgfplots}
\pgfplotsset{width=13cm,compat=1.8}
\usepackage{pgfplotstable}
\usetikzlibrary{calc}

\newtheorem{hyp}{Hypothesis}

\usepackage{color, colortbl}
\definecolor{Gray}{gray}{0.9}
\definecolor{support}{rgb}{0.6, 0.92, 0.6}
\definecolor{neutral}{rgb}{0.83, 0.83, 0.83}
\definecolor{attention}{rgb}{0.98, 1, 0.79}
\definecolor{refute}{rgb}{0.99, 0.70, 0.70}

\definecolor{true}{HTML}{1F1F1F}
\definecolor{fake}{HTML}{C2C2C2}

\usepackage{array}
\newcolumntype{P}[1]{>{\centering\arraybackslash}p{#1}}

\usepackage{booktabs}

\title{\textsf{Multiverse}: \\ Multilingual Evidence for Fake News Detection
\thanks{\textit{Extended version of} \cite{dementieva-panchenko-2021-cross}} 
}

\author{
    Daryna Dementieva \\
    Technical University of Munich \\
    Munich, Germany \\
    \texttt{daryna.dementieva@tum.de} \\
    \And
    Mikhail Kuimov, Alexander Panchenko \\
    Skolkovo Institute of Science and Technology \\
    Moscow, Russia \\
    \texttt{\{mikhail.kuimov, a.panchenko\}@skoltech.ru}
}

\begin{document}
\maketitle

\begin{abstract}
  Misleading information spreads on the Internet at an incredible speed, which can lead to irreparable consequences in some cases. It is becoming essential to develop fake news detection technologies. While substantial work has been done in this direction, one of the limitations of the current approaches is that these models are focused only on one language and do not use \textit{multilingual information}. 
  In this work, we propose \textsf{Multiverse} -- a new feature based on multilingual evidence that can be used for fake news detection and improve existing approaches. The hypothesis of the usage of cross-lingual evidence as a feature for fake news detection is confirmed, firstly, by manual experiment based on a set of known true and fake news. After that, we compared our fake news classification system based on the proposed feature with several baselines on two multi-domain datasets of general-topic news and one fake COVID-19 news dataset showing that in additional combination with linguistic features it yields significant improvements.
\end{abstract}

% keywords can be removed

\section{Introduction}
After the manipulation of opinions on Facebook during the 2016 U.S. election \cite{allcott2017social}, the interest in the topic of fake news has increased substantially. Unfortunately, the distribution of fakes leads not only to misinformation of readers but also to more severe consequences. There was a case when the spread of rumor about Hillary Clinton leading a child sex trafficking lead to Washington Pizzeria \cite{kang2016washington}. Moreover, due to the global pandemic in 2020 there was a simultaneous emergence of infodemic \cite{alamfighting} that could lead to even worse epidemiological situation and harm people's health dramatically. 

As a result, fake news received tremendous public attention, as well as drawn increasing interest from the academic community. %Especially during a increase in the amount of Internet content, special hopes are placed on automated fake news detection.
Multiple supervised fake news detection models were proposed based on linguistic features \cite{perez-rosas-etal-2018-automatic,patwa2020fighting}; deep learning models \cite{barron2019proppy,glazkova2020g2tmn,kaliyar2021fakebert,gundapu2021transformer}; or signals from social networks \cite{nguyen2020fang,shu2019defend}. One of the directions of the supervised approaches is to use additional information from the Web \cite{popat2017truth,karadzhov2017fully,ghanem2018upv}. However, in these works only monolingual signals were taken into account.

In our work, we assume that viral spreading of (fake) information may naturally hit the ``language barrier'' and cross-checking of facts across medias in various languages (supposed to be strongly independent) could yield an additional signal.  We aim to close this gap and perform an exploration of cross-lingual Web features to fake news detection.

The contributions of our work are the follows:
\begin{itemize}
    % \item We proposed new \textbf{cross-lingual evidence feature} for fake news detection based on multilingual news verification.
    \item \textsf{Multiverse}: the new \textbf{cross-lingual evidence feature} for fake news detection based on multilingual news verification is proposed.
    \item We conducted manual experiment based on cross-lingual dataset markup to evaluate if the user can use such feature for misinformation identification.
    \item We explored several strategies for cross-lingual content similarity estimation.
    \item We compared fake news classification systems based on the proposed feature with several baselines achieving SOTA results.
    \item We investigated the best models with integrated cross-lingual feature in terms of explainability, showing the examples how extracted cross-lingual information can be used for evidence generation.
\end{itemize}

The corresponding code of the \textsf{Mutliverse} feature calculation is available online.\footnote{\href{https://github.com/s-nlp/multilingual-fake-news}{https://github.com/s-nlp/multilingual-fake-news}}

%TODO: sections.

%The contribution of our work is a new \textbf{cross-lingual evidence} feature for fake news detection based on multilingual news verification. We conduct a manual experiment based on cross-lingual dataset markup to evaluate if the user can use such a feature for misinformation identification. Besides, we compare fake news classification systems based on the proposed feature with several baselines achieving SOTA results and adding to models more explainability.

\section{Related Work}

A substantial amount of research has been done in the field of fake news detection, which includes the creation of datasets and methods. In this section, we perform a comprehensive analysis of the prior art related to the subject of this article. 

\subsection{Users Behaviour Towards Fake News Detection}
\label{sec:fakenews_users}
Firstly, before the discussion of automatic machine fake news detection methods, we want to analyze the case of how real-life users react to fake information and in which way they check the veracity of information.

In \cite{lewandowsky2012misinformation} a very broad analysis of users' behavior was obtained. The authors found out that when people try to check information credibility they rely on a limited set of features, such as:

\begin{itemize}
    \item Is this information compatible with other things I believe to be true?
    \item Is this information internally coherent? Do the pieces form a plausible story?
    \item Does it come from a credible source?
    \item Do other people believe it? 
\end{itemize}

So, people can rely on the text of the news and its source and their judgment. However, if they get enough internal motivation, they can also refer to some external sources for evidence seeking. These external sources can be some knowledge sources or other people.

The conclusions from \cite{tandoc2018audiences} repeat the previous results: individuals rely on both their judgment of the source and the message, and when this does not adequately provide a definitive answer, they turn to external resources to authenticate news. The intentional and institutional reaction was seeking confirmation from institutional sources (some respondents answered simply ``Google''). 

Also, several works have been done to explore the methods to combat received by users fake information and convince them with true facts. In \cite{ecker2017reminders} it was shown that explicitly emphasizing the myth and even its repetition with refutation help users to pay attention and memorize the truth. Moreover, participants that received messages across different media platforms \cite{zhao2019misinformation} and different perspectives of the information \cite{geeng2020fake} showed greater awareness of news evidence. Consequently, the information from the external search is an important feature for news authenticity evaluation and evidence seeking. Also, a different perspective from different media adds more confidence in the decision-making process.

%Therefore, the users rely not only on the linguistic text features but also use verification across multiple resources as found on the Web, e.g. through a search engine. 
%However, previous works did not fully use this information of fake news detection. In our study we want to explore fake news spread in Web for different languages and extend evidence retrieval to cross-lingual news verification.

\subsection{Fake News Detection Datasets}
\label{sec:fakenews_datasets}
To leverage the task of automatic fake news detection there have been created several news datasets focused on misinformation, each with a different strategy of labeling.

\textbf{The Fake News Challenge}\footnote{\href{http://www.fakenewschallenge.org}{http://www.fakenewschallenge.org}} launched in 2016 was a big step in identifying fake news. The task of FNC-1 was stance detection type task \cite{hanselowski-etal-2018-retrospective}. The dataset consists of 300 topics, with 5--20 news articles for each. In general, it consists of 50K labeled claim-article pairs. The dataset is derived from the Emergent project \cite{silverman2017emergent}.

Another publicity available dataset is \textbf{LIAR} \cite{wang2017liar}. In this dataset 12.8K manually labeled short statements in various contexts from PolitiFact.com\footnote{\href{https://www.politifact.com}{https://www.politifact.com}} were collected. They covered such topics as news releases, TV or radio interviews, campaign speeches, etc. The labels for news truthfulness are fine-grained in multiple classes: pants-fire, false, barely-true, half-true, mostly true, and true.

Claim verification is also related to Fact Extraction and VERification dataset (\textbf{FEVER}) \cite{thorne-etal-2018-fever}. 185,445 claims were manually verified against the introductory sections of Wikipedia pages and classified as \textsf{SUPPORTED}, \textsf{REFUTED}, or \textsf{NOTENOUGHINFO}. For the first two classes, the annotators also recorded the sentences forming the necessary evidence for their judgment. 

\textbf{FakeNewsNet} \cite{shu2018fakenewsnet} contains two comprehensive datasets that includes news content, social context, and dynamic information. Moreover, as opposed to all the datasets described above, in addition to all textual information, there is also a visual component saved in this dataset. All news were collected with PolitiFact and GossipCop\footnote{\url{https://www.gossipcop.com}} crawlers. In general, 187014 fake and 415645 real news were crawled.

Another collected for supervised learning dataset is \textbf{FakeNewsDataset} \cite{perez-rosas-etal-2018-automatic}. The authors did a lot of manual work to collect and verify the data. As a result, they managed to collect 240 fake and 240 legit news on six different domains -- sports, business, entertainment, politics, technology, and education. All news samples are for the 2018 year.
 
One of the latest large datasets is \textbf{NELA-GT-2018} \cite{norregaard2019nela}. In this dataset authors tried to overcome some limitations that can be observed in previous works: \textit{1) Engagement-driven} -- the majority of the datasets, both for news articles and claims, contain only data that has been highly engaged with on social media or has received attention from fact-checking organizations; \textit{2) Lack of ground truth labels} -- all of the current large-scale news article datasets do not have any form of labeling for misinformation research. To overcome these limitations, they gathered a wide variety of news sources from varying levels of veracity and scraped article data from the gathered sources’ RSS feeds twice a day for 10 months in 2018. As a result, a new dataset was created consisting of 713,534 articles from 194 news and media producers.

\begingroup
\renewcommand{\arraystretch}{1.5} % Default value: 1
\begin{table}[h!]
    \centering
    \begin{tabular}{p{6cm}|p{4.5cm}|p{3cm}}
        \toprule
        \textbf{Dataset} & \textbf{Task} & \textbf{Language} \\
        \hline
        FNC-1  \cite{hanselowski-etal-2018-retrospective} & \multirow{2}{4.5cm}{Stance Detection} & English \\ \cline{1-1} \cline{3-3}
        %\hline
        Arabic Claims Dataset \cite{hasanain2019overview} &  & Arabic \\
        \hline
        FEVER \cite{thorne-etal-2018-fever} & \multirow{2}{4.5cm}{Fact Checking} & English \\ \cline{1-1} \cline{3-3}
        %\hline
        DanFEVER \cite{norregaard-derczynski-2021-danfever} & & Danish \\
        \hline
        LIAR \cite{wang2017liar} & \multirow{7}{4.5cm}{Fake News Classification} & \multirow{4}{3cm}{English} \\ \cline{1-1}
        %\hline
        FakeNewsNET \cite{perez-rosas-etal-2018-automatic} & & \\ \cline{1-1}
        %\hline
        FakeNewsDataset \cite{perez-rosas-etal-2018-automatic} &  & \\ \cline{1-1}
        %\hline
        NELA-GT-2018 \cite{norregaard2019nela} & & \\ \cline{1-1}
        %\hline
        ReCOVery \cite{zhou2020recovery} & & \\ \cline{1-1} \cline{3-3}
        %\hline
        GermanFakeNC \cite{vogel2019fake} & & German \\ \cline{1-1} \cline{3-3}
        %\hline
        The Spanish Fake News Corpus \cite{posadas2019detection} & & Spanish \\
        \bottomrule
    \end{tabular}
    \caption{The datasets covered in related work. It can be observed that the majority of the data for different fake news detection tasks is for the English language.}
    \label{tab:my_label}
\end{table}
\endgroup

Due to the events of 2020, the work has been already done in the direction of the creation COVID-19 fake news detection dataset. \textbf{COVID-19 Fake News} \cite{patwa2020fighting} was built based on the information from public fact-verification websites and social media. It consists of 10,700 tweets (5600 real and 5100 fake posts) connected with the COVID-19 topic. In addition, there was created \textbf{ReCOVery} \cite{zhou2020recovery} multimodal dataset. It also incorporates in itself 140,820 labeled tweets as well as 2,029 news articles on coronavirus collected from reliable and unreliable resources.

However, all of the above datasets have one main limitation -- they are monolingual and dedicated only to the English language. Talking about other languages other than English, such datasets can be mentioned: \textit{French satiric dataset} \cite{liu2019detection}, \textit{GermanFakeNC} \cite{vogel2019fake}, \textit{The Spanish Fake News Corpus} \cite{posadas2019detection}, \textit{Arabic Claims Dataset} \cite{hasanain2019overview}. However, all of these datasets are monolingual as well and mostly cover fake news classification tasks missing, for instance, fact verification and evidence generation problems. There was only collected \textit{A Multilingual Cross-domain Fact Check News Dataset for COVID-19} \cite{shahifakecovid} that covers 40 languages from 105 countries (English, Spanish, French, Portuguese, Hindi languages, and others). However, this dataset is highly imbalanced. Firstly, there is a disbalance in terms of fake (4132 samples) and true (1050 samples) labels. Secondly, the number of English samples is significantly bigger than for other languages: for the top first English language, there are over 2000 samples, for the top second Spanish there are almost 1000 samples, for the top third French language there are only 250 samples, and further data size for other languages decreases dramatically. All these statistics also illustrate the difficulties of collecting multilingual fake news datasets. Consequently, the creation of a supervised dataset for each language and implementation algorithm of fake news detection for each language will be a very resource- and time-consuming task.

\subsection{Fake News Classification Methods}
\label{sec:fakenews_review_methods}
% In this section we want to consider the task of specifically fake news classification. Thus, we suppose that we have a collection of labeled documents $D = \{(d_i, l_i)\}_{i=1}^{N}$, where $l_i \in \{Legit, Fake\}$. We want to build classifier $f: d \mapsto l$ that will predict the label with respect to the document $d$. 
On the basis of previously described datasets, several solutions were created to tackle the problem of obtaining such a classifier. The feature sets used in all existing methods can be divided into two categories: 1)~\textbf{internal} features that can be obtained by different preprocessing strategies and linguistic analysis of the input text; 2)~\textbf{external} features that are extracted from some knowledge base, the Internet or social networks and give additional information about the facts from the news, its propagation in social media and users reactions. 

\subsubsection{Methods based on Internal Features}

One of the types of features that are helpful in fake news classification tasks is linguistic and psycholinguistic features. In \cite{perez-rosas-etal-2018-automatic} a strong baseline model based on such a feature set was created based on the FakeNewsDataset. The feature set used in this work looks as follows:
\begin{itemize}
    \item \textbf{Ngrams}: tf-idf values of unigrams and bigrams from a bag-of-words representation of the input text.
    \item \textbf{Punctuation} such as periods, commas, dashes, question marks, and exclamation marks. 
    \item \textbf{Psycholinguistic features} extracted with LIWC lexicon. Alongside some statistical information, LIWC also provides emotional and psychological analysis.
    \item \textbf{Readability} that estimates the complexity of a text. The authors use content features such as number of characters, complex words, long words, number of syllables, word types, and others. In addition, they used several readability metrics, including the Flesch-Kincaid, Flesch Reading Ease, Gunning Fog, and Automatic Readability Index.
    \item \textbf{Syntax}: a set of features derived from production rules based on context-free grammar (CFG) trees.
\end{itemize}

Based on such features, different statistical machine learning models can be trained. In \cite{perez-rosas-etal-2018-automatic} the authors trained the SVM classifier according to the set of characteristics presented. Naïve Bayes, Random Forest, KNN, and AdaBoost were also frequently used as fake news classification models \cite{choudhary2021linguistic,sharma2019combating,gravanis2019behind}.

In \cite{ghanem2020emotional} the perspective of the usage of emotional signals extracted from the news text for detecting fakes was shown. The authors analyzed the set of emotions that are present in true and fake news checking the hypothesis that trusted news does not use emotions to affect the reader's opinion while the fake one does. They found out that such emotions as \textit{negative emotions}, \textit{disgust}, \textit{surprise} have more tendency to appear in fake news and can give a strong signal for fake news classification. 

Additionally to linguistic features, feature extraction strategies based on deep learning architectures were also explored. In \cite{kaliyar2020fndnet} the classical architecture for text classification task based on CNN was successfully applied for the fake news detection task. With the recent growth of the usage of Transformer architectures in the NLP field, such models as BERT \cite{kaliyar2021fakebert,jwa2019exbake} and RoBERTa \cite{glazkova2020g2tmn} also demonstrated high results for general-topic fakes classification as well as COVID-19 fake news detection task.

As it can be seen, one of the main advantages of models based on internal feature sets is that such models are quite easy to use and they do not require significant additional time for feature extraction. Moreover, such models can be optimal in terms of inference time and memory usage because they only operate with internal information from input news. However, if we consider the explainability aspect for the end users, the evidence generated from such internal features most likely will be not enough to convince the user of the correctness of model performance and to motivate the label decision for the news. 

\subsubsection{Methods based on External Non-Textual Features}
Although internal features-based models can achieve high classification scores in the fake news classification task, the decision of such is hard to interpret. As a result, additional signals from external sources can add more confidence to model decision reasoning.

If the news appears in some social network, the information about the users that liked or reposted the news post and the resulted post propagation can be used as a feature for fake news classification. It was shown in \cite{zhao2020fake} that fake news spread over social networks quicker after the publication than true news. As a result, to combat fake news in the early stages of its appearance, several methods have been created to detect the anomaly behavior in reposts or retweets \cite{liu2018early,shu2019beyond}. In \cite{shu2019role} the different information about specific users was explored. The author extracted location, profile image, and political bias to create a feature set.

Another type of information that can be obtained from users and be used as some kind of knowledge base is users' comments related to the news post. This approach was explored in \cite{shu2019defend}. There was created \textit{dEFEND} system for explainable fake news detection. The information from users' comments was used to find related evidence and validate the fact from the original news. \textit{Factual News Graph (FANG)} system from \cite{nguyen2020fang} was presented to connect the content of news, news sources, and user interaction to build a full-filled social picture about the inspected news.

Talking about the information verification step in the fake news detection pipeline, there were created several methods for leveraging a fact-checking task. One of the sources for providing a knowledge base with evidence is Wikipedia. The FEVER dataset that was previously discussed in Section \ref{sec:fakenews_datasets} consists of claims and evidence already pre-extracted from Wikipedia. Several works like \cite{soleimani2020bert,atanasova2020generating,nie2019combining} are dedicated to the fact-checking task and evidence generation based on Wikipedia pages.

On the other hand, the knowledge base for obtaining evidence for information verification can be simply the Web. In \cite{popat2017truth,karadzhov2017fully,ghanem2018upv,li2020connecting} the authors referred to the Web search (Google or Bing) to collect relevant articles and use such scraped information as an external feature to build fake news classifier. As it was discussed in Section \ref{sec:fakenews_users}, such a Web-based feature is quite motivated by real-life users' behavior. As a result, the generated evidence based on the Web scraped information can be more persuasive for the users as it automatizes the steps that they take to check the veracity of the news.

However, in all the discussed methods we can also see the usage of only one language for evidence granting. The systems that used Web search for evidence extraction turned to only English search results. In our work, we want to fill this gap to explore cross-lingual Web-based evidence for the fake news classification task. 

\section[\textsf{Multiverse}: A New Feature for Fake News \\ Classification]{\textsf{Multiverse}: A New Feature for Fake News Classification}
% \sectionmark{\textsf{Multiverse}}
% \markright{\textsf{Multiverse}}
\label{sec:fakenews_method}
We present \textsf{Multiverse} -- \underline{Multi}lingual E\underline{v}idenc\underline{e} fo\underline{r} Fake New\underline{s} D\underline{e}tection based on extraction from Web search.
% We present a new method for fake news detection based on \textbf{cross-lingual evidence (CE)} extraction from Web search.
The idea is motivated by the user experience illustrated in Section \ref{sec:fakenews_users} and the lack of multilingualism in automatic fake news detection methods, as discussed in Section \ref{sec:fakenews_review_methods}. Users quite often refer to the Web search to check news seen in some news feed. However, to show the different points of view and additional information out of a monolingual bubble, the cross-lingual check of original news can be quite persuasive and can give a larger room for rational judgment about information.

\begin{figure*}[h!]
  \centering
  \includegraphics[width=1.0\textwidth]{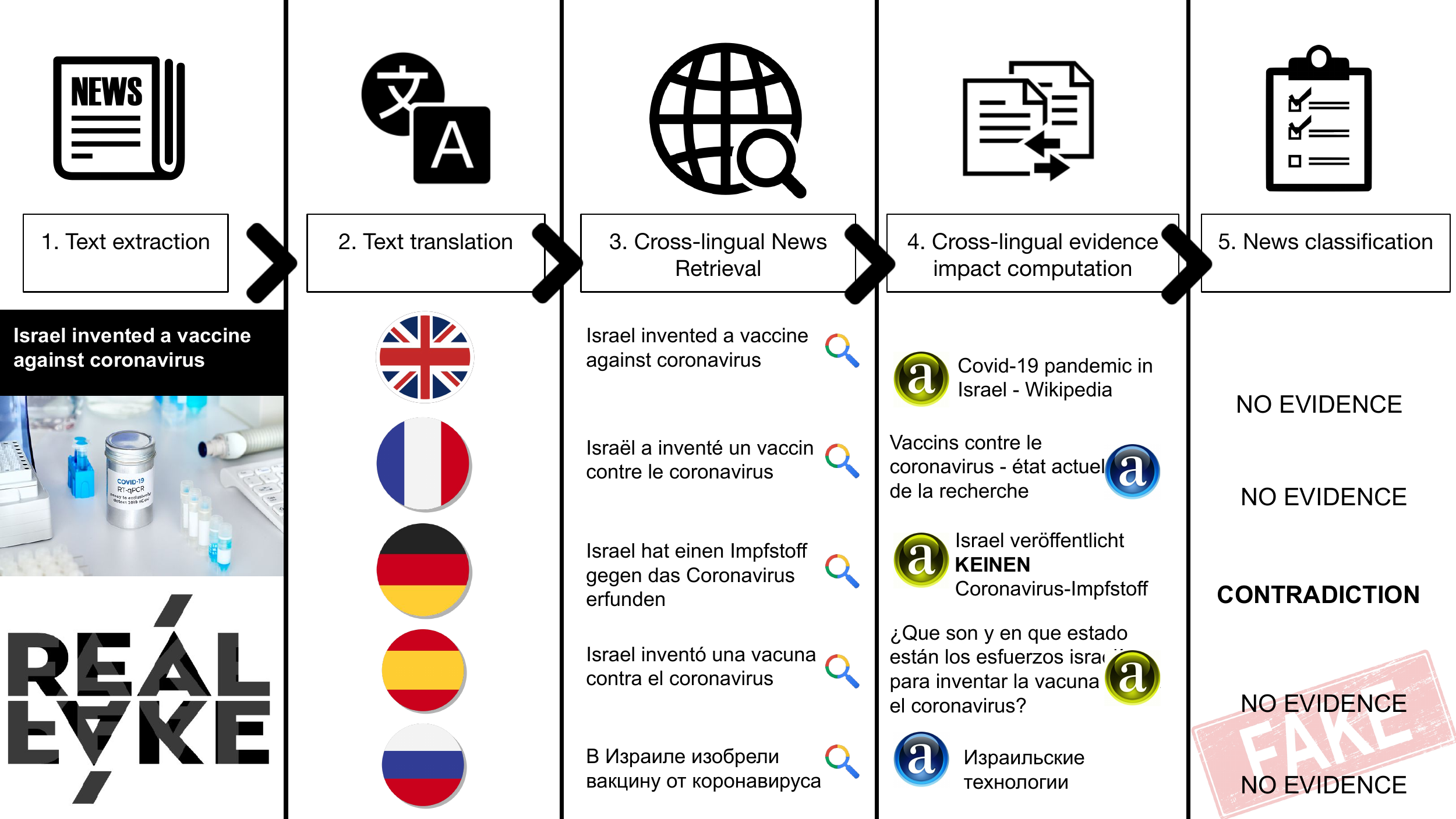}
  \caption{Overview of our approach: checking for fake news based on cross-lingual evidence (CE).}
  \label{fig:schema}
\end{figure*}

Our proposed approach is based on the following hypothesis: 
\begin{hyp}[H\ref{hyp:first}] \label{hyp:first}
\leavevmode
\begin{itemize}
    \item If the news is \underline{true}, then it will be widespread in different languages and also across media with different biases, and the facts mentioned should be identical.
    \item If the news is \underline{fake}, it will receive a lower response in the foreign press than a piece of true news. 
\end{itemize}
\end{hyp}

The step-by-step pipeline of the approach, schematically represented in Figure \ref{fig:schema}, is as follows:

\begin{itemize}

\item \textbf{Step 1. Text extraction:} As a new article arrives, the title and content are extracted from it.

\item \textbf{Step 2. Text translation:} The title is translated into target languages and new search requests are generated. 

\item \textbf{Step 3. Cross-lingual news retrieval:} Based on generated cross-lingual request -- translated title -- the search with a Web search engine is executed.

\item \textbf{Step 4. Cross-lingual evidence impact computation:} Top-N articles from search results are extracted to assess the authenticity of the initial news. The information described in the news is compared with the information in the articles from the search result. Also, the ranks of the source of the extracted articles are taken into account. The number of articles that confirms or disproves the original news from reliable sources is estimated.

\item \textbf{Step 5. News classification:} Based on the information from the previous step, the decision is made about the authenticity of the news. If the majority of results support the original news, then it is more likely to be true; if there are contradictions -- it is a signal to consider the news as a fake.

\end{itemize}

As we can see from the example in the scheme in Figure \ref{fig:schema}, for the news \textit{``Israel invented a vaccine against coronavirus"} the majority of the scraped articles provide no evidence that supported incoming news. Moreover, there is an article with high reliability that provides an explicit refutation of the original information. As there is none of the supporting information and a contradiction with the scraped information, the probability that we should believe in the veracity of this news is quite low.

The proposed method based on cross-lingual evidence extraction can work properly with worldwide important news. Indeed, if there is some local event about locally famous parties, in the majority of cases such news will be doubtfully widespread all over the Internet. As a result, in our future assumptions and experiments, we take into consideration datasets and news that cover worldwide events.

To incorporate the proposed feature into an automatic fake new detection pipeline, firstly, we wanted to lean on user experience and check the following hypothesis:
\begin{hyp}[H\ref{hyp:second}] \label{hyp:second}
The person can detect fake news using cross-lingual evidence using the pipeline presented in Figure \ref{fig:schema}.
\end{hyp}

After this hypothesis confirmation, we can explore the possibilities to automate fake news classification using the cross-lingual evidence feature confirming the next hypothesis:
\begin{hyp}[H\ref{hyp:third}] \label{hyp:third}
The proposed cross-lingual evidence feature can improve automatic fake news detection.
\end{hyp}

\begin{table*}[h!]
    \centering
    \scriptsize
    \begin{tabular}{p{7.25cm}|p{6.25cm}|c}
    \toprule
    \textbf{News title} & \textbf{URL} & \textbf{Label} \\
    \hline
    Lottery winner arrested for dumping \$200,000 of manure on ex-boss’ lawn &    \href{https://worldnewsdailyreport.com/lottery-winner-arrested-for-dumping-200000-of-manure-on-ex-boss-lawn/}{https://worldnewsdailyreport.com/lottery-winner-arrested-for-dumping-200000-of-manure-on-ex-boss-lawn/} & \color{red}{Fake} \\ 
    \hline
    % please include (duplicate the real url instead of google.com): masterclass how to hack the system
    Woman sues Samsung for \$1.8M after cell phone gets stuck inside her vagina & \href{https://worldnewsdailyreport.com/woman-sues-samsung-for-1-8m-after-cell-phone-gets-stuck-inside-her-vagina/comment-page-58/}{https://worldnewsdailyreport.com/woman-sues-samsung-for-1-8m-after-cell-phone-gets-stuck-inside-her-vagina/comment-page-58/} & \color{red}{Fake} \\
    \hline
    BREAKING: Michael Jordan Resigns From The Board At Nike-Takes 'Air Jordans' With Him & \href{https://www.newsbreak.com/news/944830700924/breaking-michael-jordan-resigns-from-the-board-at-nike-takes-air-jordans-with-him}{https://www.newsbreak.com/news/944830700924\newline/breaking-michael-jordan-resigns-from-the-board-at-nike-takes-air-jordans-with-him} & \color{red}{Fake} \\
    \hline
    Donald Trump Ends School Shootings By Banning Schools & 
    \href{https://www.8shit.net/donald-trump-ends-school-shootings-banning-schools/}{https://www.8shit.net/donald-trump-ends-school-shootings-banning-schools/} & \color{red}{Fake} \\
    \hline
    New mosquito species discovered that can get you pregnant with a single bite &
    \href{https://thereisnews.com/new-mosquito-species-discovered-can-make-you-pregnant/}{https://thereisnews.com/new-mosquito-species-discovered-can-make-you-pregnant/} & \color{red}{Fake} \\
    \hline
    Obama Announces Bid To Become UN Secretary General &
    \href{https://www.pinterest.com/pin/465630048969491948/}{https://www.pinterest.com/pin/46563004896949\newline1948/} & \color{red}{Fake} \\
    \hline
    Lil Tay Rushed To Hospital After Being Beat By Group Of Children At A Playground &
    \href{https://www.huzlers.com/lil-tay-rushed-to-hospital-after-being-beat-by-group-of-children-at-a-playground/}{https://www.huzlers.com/lil-tay-rushed-to-hospital-after-being-beat-by-group-of-children-at-a-playground/} & \color{red}{Fake} \\
    \hline
    Post Malone's Tour Manager Quits Says Post Malone Smells Like Expired Milk And Moldy Cheese &     \href{https://www.huzlers.com/post-malones-tour-manager-quits-says-post-malone-smells-like-expired-milk-and-moldy-cheese/}{https://www.huzlers.com/post-malones-tour-manager-quits-says-post-malone-smells-like-expired-milk-and-moldy-cheese/} & \color{red}{Fake} \\
    \hline
    Putin: Clinton Illegally Accepted \$400 Million From Russia During Election & 
    \href{https://newspunch.com/putin-clinton-campaign-400-million-russia/}{https://newspunch.com/putin-clinton-campaign-400-million-russia/} & \color{red}{Fake} \\
    \hline
    Elon Musk: 99.9\% Of Media Is Owned By The 'New World Order' &
    \href{https://newspunch.com/elon-musk-media-owned-new-world-order/}{https://newspunch.com/elon-musk-media-owned-new-world-order/} & \color{red}{Fake} \\ 
    \midrule 
    Scientists Develop New Method to Create Stem Cells Without Killing Human Embryos &
    \href{https://www.christianpost.com/news/scientists-develop-new-method-to-create-stem-cells-without-killing-human-embryos.html}{https://www.christianpost.com/news/scientists-develop-new-method-to-create-stem-cells-without-killing-human-embryos.html} & \color{blue}{Legit} \\
    \hline
    Luis Palau Diagnosed With Stage 4 Lung Cancer &   \href{https://cnnw.com/luis-palau-diagnosed-with-stage-4-lung-cancer/}{https://cnnw.com/luis-palau-diagnosed-with-stage-4-lung-cancer/} & \color{blue}{Legit} \\
    \hline
    1st black woman nominated to be Marine brigadier general &   \href{https://edition.cnn.com/2018/04/12/politics/marine-corps-brigadier-general-first-black-female/index.html}{https://edition.cnn.com/2018/04/12/politics/\newline marine-corps-brigadier-general-first-black-female/index.html} & \color{blue}{Legit} \\
    \hline
    Disney CEO Bob Iger revealed that he seriously explored running for president &   \href{https://www.businessinsider.com/disney-ceo-bob-iger-says-he-considered-running-for-president-oprah-pushed-2018-4}{https://www.businessinsider.com/disney-ceo-bob-iger-says-he-considered-running-for-president-oprah-pushed-2018-4} & \color{blue}{Legit} \\
    \hline
    Trump Has Canceled Via Twitter His G20 Meeting With Vladimir Putin &   \href{https://www.buzzfeednews.com/article/emilytamkin/trump-g20-putin-russia}{https://www.buzzfeednews.com/article/emily\newline tamkin/trump-g20-putin-russia} &  \color{blue}{Legit} \\
    \hline
    US Mexico and Canada sign new USMCA trade deal &   \href{https://www.dw.com/en/us-mexico-canada-sign-usmca-trade-deal/a-51613992}{https://www.dw.com/en/us-mexico-canada-sign-usmca-trade-deal/a-51613992} & \color{blue}{Legit} \\
    \hline
    Afghanistan Women children among 23 killed in US attack UN &   \href{https://www.aljazeera.com/news/2018/11/30/afghanistan-women-children-among-23-killed-in-us-attack-un}{https://www.aljazeera.com/news/2018/11/30/\newline afghanistan-women-children-among-23-killed-in-us-attack-un} & \color{blue}{Legit} \\
    \hline
    UNESCO adds reggae music to global cultural heritage list &   \href{https://www.aljazeera.com/features/2018/11/29/unesco-adds-reggae-music-to-global-cultural-heritage-list}{https://www.aljazeera.com/features/2018/11/29/\newline unesco-adds-reggae-music-to-global-cultural-heritage-list} & \color{blue}{Legit} \\
    \hline
    The Saudi women detained for demanding basic human rights &   \href{https://www.aljazeera.com/news/2018/11/29/the-saudi-women-detained-for-demanding-basic-human-rights/}{https://www.aljazeera.com/news/2018/11/29/the-saudi-women-detained-for-demanding-basic-human-rights/} & \color{blue}{Legit} \\
    \hline
    Georgia ruling party candidate Zurabishvili wins presidential runoff &   \href{https://www.aljazeera.com/news/2018/11/30/ex-envoy-wins-georgia-presidency-vote-to-be-challenged}{https://www.aljazeera.com/news/2018/11/30/ex-envoy-wins-georgia-presidency-vote-to-be-challenged} & \color{blue}{Legit} \\
    \bottomrule
    \end{tabular}
    \caption{The manually selected 20 news dataset (10 fake and 10 true news) for manual experiment. Fake news were selected from the top 50 fake news of 2018 according to BuzzFeed. Legit news were selected from NELA-GT-2018 dataset.}
    \label{tab:fakenews_20newsdata}
\end{table*}

% \section{Evaluation}
To confirm all the above hypotheses we conducted several experiments. For all experiments, we chose top-5 European languages spoken in Europe\footnote{\href{https://www.justlearn.com/blog/languages-spoken-in-europe}{https://www.justlearn.com/blog/languages-spoken-in-europe}} and used in Internet\footnote{\href{https://www.statista.com/statistics/262946/share-of-the-most-common-languages-on-the-internet}{https://www.statista.com/statistics/262946/share-of-the-most-common-languages\\-on-the-internet}} -- English, French, German, Spanish, and Russian -- to obtain cross-lingual evidence. For the search engine, we stopped at Google search\footnote{\href{https://www.google.com}{https://www.google.com}} as it is the top-1 search engine in the world\footnote{\href{https://www.oberlo.com/blog/top-search-engines-world}{https://www.oberlo.com/blog/top-search-engines-world}} and also claimed to be widely used by users during use case fake news check experiment mentioned in Section \ref{sec:fakenews_users}.

The first experiment is a manual small-scale study confirming Hypothesis \ref{hyp:first} and Hypothesis \ref{hyp:second}. After that, we tested several approaches to automatize the pipeline and compared them with manual markup (Section~\ref{sec:fakenews_manual_evaluation}). The final step (Section~\ref{sec:fakenews_second_experiment}) of the confirmation of Hypothesis \ref{hyp:third} is an automated fake news detection system tested on several fake news datasets: we implemented our cross-lingual evidence feature and compared it with several baselines achieving SOTA on all datasets.

\section{Experiment 1: Manual Verification}
\label{sec:fakenews_manual_evaluation}

To confirm Hypothesis \ref{hyp:first} and Hypothesis \ref{hyp:second} we conducted an experiment with manual markup where the annotators were asked to classify fake news based on cross-lingual evidence.

\subsection{Dataset}
For fake news examples, we used the list of top 50 fake news from 2018 according to BuzzFeed.\footnote{\href{https://github.com/BuzzFeedNews/2018-12-fake-news-top-50}{https://github.com/BuzzFeedNews/2018-12-fake-news-top-50}}.
For true news, we used NELA-GT-2018 dataset \cite{norregaard2019nela}. We manually selected 10 fake and true news. We tried to cover several topics in this dataset: celebrities, science, politics, culture, and the world. The full dataset featuring 20 news used for the manual markup is provided in Table \ref{tab:fakenews_20newsdata}.

\subsection{Experimental Setup}

\begin{figure*}[t!]
    \centering
        \includegraphics[width=\textwidth]{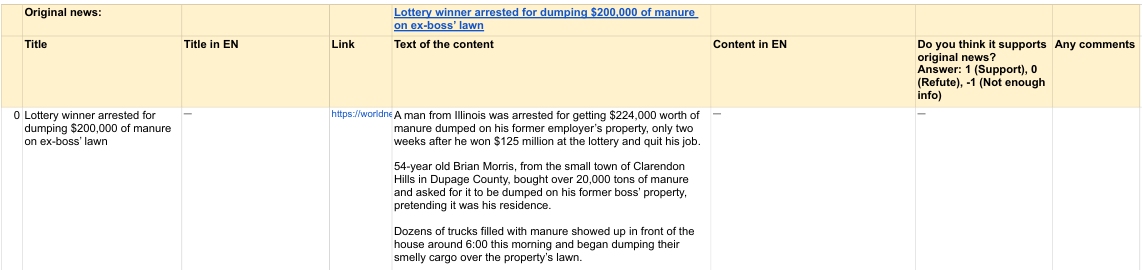}\\
        \vspace{0.5cm}
        \includegraphics[width=\textwidth]{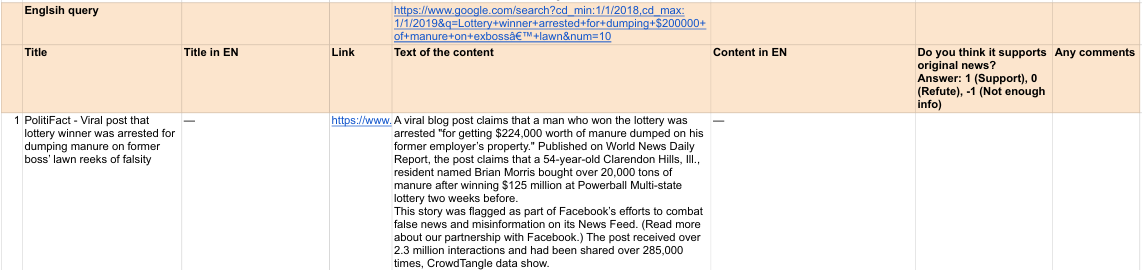}\\
        \vspace{0.5cm}
        \includegraphics[width=\textwidth]{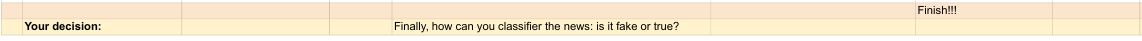}
    \caption{User interface that was used for annotators answer collection for manual verification. The annotator was provided with original news and the link to the source. After that he was given the results of cross-lingual search results with translation into English if needed. For each news from search result the title, link to the source, and text of the content were provided. The task of the annotator was to identify if the scraped news supported, refuted the original news or provided not enough information to make a decision. As a final step, the annotator was asked to do the classification of the original news into fake or true.}
    \label{fig:manual_interface}
\end{figure*}

% For cross-lingual information extraction setup (\textbf{Step 2}) we chose top-5 languages spoken in Europe\footnote{https://www.justlearn.com/blog/languages-spoken-in-europe}: English (original language of news), French, German, Spanish, and Russian.

%For each news from the dataset we pre-conducted translation of the titles and cross-lingual search (\textbf{Step 3}) for annotators convenience (and future reproducibility). For the translation and search the Google services were used. As the news are of 2018, the time range of every search was limited only by this year. From search results we used the first page of the search which consisted of 10 news. As a result, for 20 news for each of languages we got 1000 pairs of ``original news $\leftrightarrow$ scraped news'' to markup.

As nowadays Google provides personalized search results\footnote{\href{http://googlepress.blogspot.com/2004/03/google-introduces-personalized-search.html}{http://googlepress.blogspot.com/2004/03/google-introduces-personalized-search.html}}, we precalculated \textbf{Step 2} and \textbf{Step 3} for annotators convenience and reproducibility. We generated cross-lingual requests in five languages -- English, French, German, Spanish, and Russian. For translation from English, the Google Translation service was used. As the news are of 2018, the time range of every search was limited only to this year. For the cross-lingual search, the translated titles were used. From search results, we used the first page of the search which consisted of 10 news. As a result, for 20 news for each of all languages we got 1000 pairs of ``original news $\leftrightarrow$ scraped news'' to markup. 

We asked 6 annotators to take part in the experiment: manually conduct \textbf{Step 4}: cross-lingual evidence impact computation. For this, we created an interface for the markup presented in Figure \ref{fig:fakenews_manual_experiment}. For each piece of news, we provide information about its title, content, and link to the source. As a result, every annotator could evaluate the quality of the text, the credibility of the source, and cross-lingual evidence for each sample from the dataset.

Every annotator got 10 randomly selected news, as a result, we got each news cross-checked by 3 annotators. All non-English pieces of news were translated into English. For each pair ``original news $\leftrightarrow$ scraped news'' the annotator provided one of three answers: 1) \textbf{Support}: the information in the scraped news supports the original news; 2) \textbf{Refute}: the information is opposite or differ from the original news or there is an explicit refutation; 3) \textbf{Not enough info}: the information is not relevant or not sufficient to support/refute the original news. Finally, at the end of the annotation of a news, the annotator was asked to conduct \textbf{Step 5} of the pipeline and classify the news as fake or true.

\subsection{Discussion of Results}

\begin{figure*}[ht!]
\hspace*{-0.25cm}
    \centering
    \includegraphics[scale=0.325]{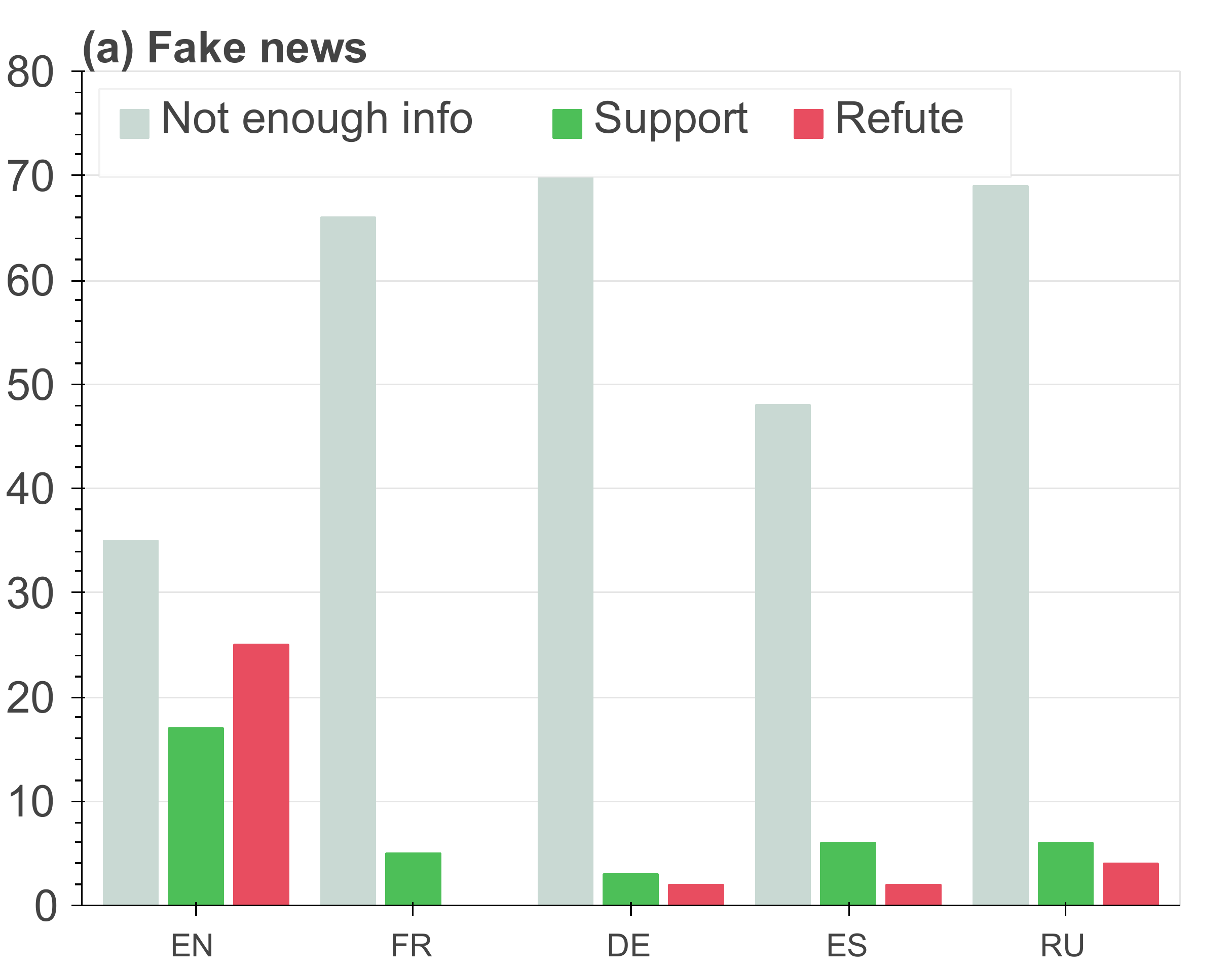}
\hspace*{-0.15cm}
    \includegraphics[scale=0.325]{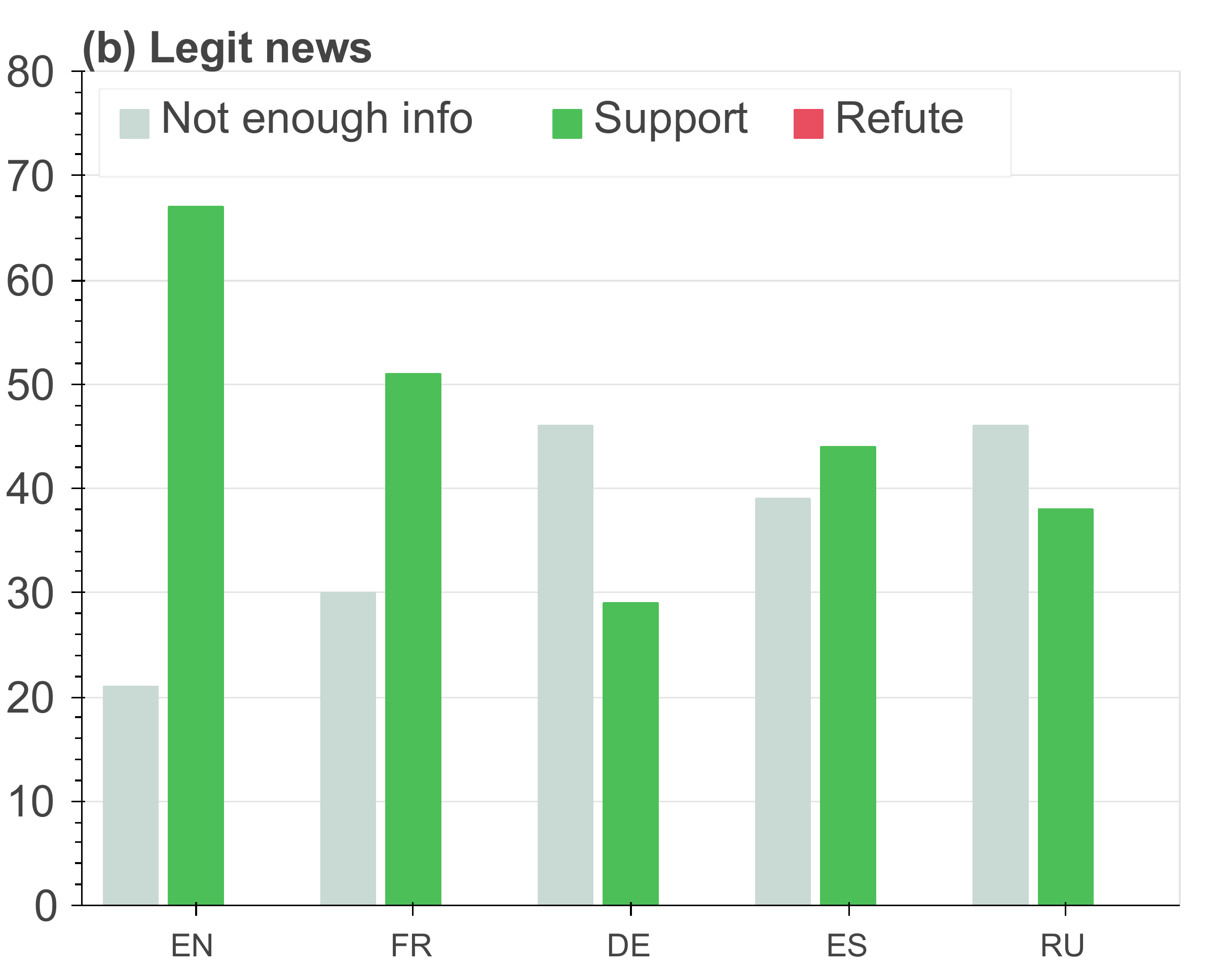}
    \caption{The results of manual annotation: the distribution of annotators answers for fake (a) and legit (b) news. As we can see, the amount of Support news from search results for every language for legit news incredibly overcome the amount for fake news. At the same time, there is almost none of Refute news for legit news while Refute news appeared in the search results for fake news across all languages.}
    \label{fig:fakenews_manual_experiment}
\end{figure*}

Based on the collected annotations, for each news, we chose the final label based on the majority voted. We estimated confidence in the annotators' agreement with Krippendorff's alpha ($\alpha=0.83$). After that, we calculated the distribution of each type of annotator's answers for the top 10 search results by languages for fake and true news separately. The results are provided in Figure \ref{fig:fakenews_manual_experiment}. 

As we can see, the distribution of labels for true news significantly differs from the distribution for fake ones: the number of supporting articles is enough for almost every language. At the same time, for fake news, we got more refuting signals than supporting the English language and little or no evidence or relevant information dissemination for other languages. The obtained result can be used for Hypothesis~\ref{hyp:first} confirmation: the fake news indeed received less spread over different languages, while for true news we can see supportive information from multilingual sources. Finally, the average accuracy of annotators classification is $0.95$. That confirms our Hypothesis \ref{hyp:second}: a person can distinguish fake news based on cross-lingual evidence.

\section{Experiment 2: Automatic Verification}
\label{sec:fakenews_second_experiment}

After the manual verification of the proposed feature, we conducted the chain experiments to validate Hypothesis \ref{hyp:third}. To achieve that, we automated all the steps of the pipeline presented in Section \ref{sec:fakenews_method}. We experimented with several approaches for cross-lingual evidence feature computation and compared the implementations with annotators markup obtained in Section \ref{sec:fakenews_manual_evaluation}. After that, we incorporated our feature in an automated fake news detection pipeline comparing with baseline methods.
%\subsection{Automatic Approach}
%\section{Experiment 2: Automatic Verification}
\subsection{Automatic Cross-lingual Evidence Feature}
\label{sec:experiment2}
Firstly, we implemented the cross-lingual evidence feature according to the steps of the pipeline described in Section \ref{sec:fakenews_method}. We implemented Algorithm~\ref{alg:multiverse} that automatically extracts cross-lingual evidence features for input news. 

\begin{algorithm}[h!]
\caption{Multilingual Evidence for Fake News detection: feature extraction. \\
\textit{\textbf{Input}}: news information $n$, languages to use for comparison $l \in L$ the maximum amount of news from Web search to compare with $N$ \\
\textit{\textbf{Output}}: cross-lingual evidence feature set $(s_i,a_i)$ of similarity with the original news and source credibility rank for each news $w_i$ from multilingual web search.
}
\label{alg:multiverse}
\begin{algorithmic}[1]
\Function{cosine\_distance\_news\_similarity}{$n, w, l$} 
\If{$type(w) \operatorname{is not} text$}
    \State $news\_pair\_similarity$ = 0
\EndIf

\If{$[l(\operatorname{``fake"}), l(\operatorname{``false"}), l(\operatorname{``lie"})] \in w$}
    \State $news\_pair\_similarity$ = 0
\EndIf

\State $news\_pair\_similarity = \operatorname{cosine\_distance}(\operatorname{mBERT}(n), \operatorname{mBERT}(w))$

\State \Return $news\_pair\_similarity$
\EndFunction
\\\hrulefill
\Function{nli\_news\_similarity}{$n, w, l$} 

\State $news\_pair\_similarity = \operatorname{XNLI-RoBERTa}(n, w)$

\State \Return $news\_pair\_similarity$
\EndFunction
\\\hrulefill
\end{algorithmic}

\begin{algorithmic}[1]
\Function{Multiverse}{$n$, $L$, $N$}

\State $cross\_lingual\_evidence \coloneqq []$

\For{$l \in L$}
    \State $headline_l = \operatorname{Translate}(n[headline], lang=l)$
    \State $W = \operatorname{Search}(headline_l, top=N)$
    \For{$w \in W$}
        \State $source\_rank = \operatorname{AlexaRank}(w)$
        \State \# For similarity score cosine- or nli-based function can be chosen
        \State $similarity = \operatorname{cross\_lingual\_news\_similarity(n, w, l)}$
        \State $cross\_lingual\_evidence.\operatorname{append}(similarity, source\_rank)$
    \EndFor
\EndFor

\State \Return $cross\_lingual\_evidence$

\EndFunction
\end{algorithmic}

\end{algorithm}

\subsubsection{Cross-lingual evidence retrieval}
% As in manual verification setup, for translation and search we used Google services via Python APIs. In our setup for the automated feature we focused as well on five languages: English, French, German, Spanish, and Russian.

    % \item Translate news' title into target languages with Googletrans\footnote{\url{https://pypi.org/project/googletrans}}.
    % \item Execute a search on new requests with Google Search API\footnote{\url{https://pypi.org/project/Google-Search-API}}. 
    % We took into account the first page of a search result.
    % \item Extract titles and contents of search results with Beautiful Soup\footnote{\url{https://pypi.org/project/beautifulsoup4}}.
    
To automate \textbf{Step 2}: \textit{Text translation}, we used Googletrans\footnote{\href{https://pypi.org/project/googletrans}{https://pypi.org/project/googletrans}} library. For the translation, we used five languages as well:  English, French, German, Spanish, and Russian. To execute \textbf{Step 3}: \textit{Cross-lingual News Retrieval}, the Google Search API\footnote{\href{https://pypi.org/project/Google-Search-API}{https://pypi.org/project/Google-Search-API}} was used. As in the manual experiment, we generated the queries as the translated titles of the original news and extracted only the first page of the search result which gave us 10 articles for each language.

\subsubsection{Content similarity computation}
\label{sec:fakenews_content_similarity_baseline}
The goal of \textbf{Step 4}: \textit{Cross-lingual evidence impact computation} is to figure out if the information in scraped articles supports or refutes the information from the original news. To compute this measurement we tested two strategies: 1) similarity computation based on cosine distance between text embeddings; 2) scores based on NLI model.

\paragraph{Cosine distance} Firstly, we evaluated the similarity between two news based on their texts' embeddings. As the similarity between text embeddings can be interpreted as the similarity between text content, we assumed that such a strategy for content similarity computation can correlate with the fact that one news support information from another one. However, there can be cases when the contents of the news can be very close or even duplicated, but the special remarks such as ``Fake", ``Rumor", etc. indicate the refutation of the original facts. We took into account such situations. As a result, the algorithm for this approach of content similarity computation looks as follows:
\begin{enumerate}
    \item If the link from the search leads to the file and not to the HTML page, then the news at this link is automatically considered dissimilar to the original one;
    \item If there are signs of disproof of news such as the words ``fake", ``false", ``rumor", ``lie" (and their translations to the corresponding language), negations, or rebuttal, then the news is automatically considered dissimilar to the original one;
    \item Finally, we calculate the similarity between the news' title and the translated original one. For a similarity measure, we choose cosine similarity between sentence embeddings. To get sentence vector representation we average sentence's tokens' embeddings extracted from Multilingual Bert (mBERT) released by \cite{devlin-etal-2019-bert}. If the similarity measure overcomes the threshold $\theta$, then the information described in scraped news and original news is considered the same.
\end{enumerate}

\paragraph{Natural Language Inference (NLI)} On the other hand, the task of estimating similarity between news contents can be reformulated as Natural Language Inference task. NLI is the problem of determining whether a natural language hypothesis $h$ can reasonably be inferred from a natural language premise $p$ \cite{maccartney2009natural}. The relations between hypothesis and premise can be \textit{entailment}, \textit{contradiction} and \textit{neutral}. The release of the large NLI dataset \cite{DBLP:conf/emnlp/BowmanAPM15} and later multilingual XNLI dataset \cite{DBLP:conf/emnlp/ConneauRLWBSS18} made possible the development of different deep learning system to solve this task.

\begin{table*}[h!]
\centering
\footnotesize
\begin{tabular}{p{7cm}|p{6.75cm}|c}
    \toprule
    Premise $p$ & Hypothesis $h$ & Label  \\
    \hline
    Israel invented a vaccine against coronavirus & Israel is not releasing a coronavirus vaccine -- The Forward & \textcolor{red}{contradiction}\\
    \hline
    Israel invented a vaccine against coronavirus & Covid-19 pandemic in Israel -- Wikipedia & neutral\\
    \hline
    Israel invented a vaccine against coronavirus & Israel's vaccine has 90\% efficacy in trial & \textcolor{green}{entailment}\\
    \bottomrule
\end{tabular}
    \caption{Example how NLI model can be used to extract relations between news.}
    \label{tab:fakenews_nli_example}
\end{table*}

The number of classes and their meaning of them in the NLI task is very similar to the labels ``Support", ``Refute" and ``Not enough info" that are used for the stance detection task in the fake news detection pipeline and that we used in the manual markup. Moreover, in \cite{sadeghifake} the usage of NLI features for stance detection task based was tested. The best model based on NLI features showed a 10\% improvement in accuracy over baselines on the FNC-1 dataset. The example of the usage of the NLI model on news titles is presented in Table \ref{tab:fakenews_nli_example}.

We used XLM-RoBERTa-large model pretrained on multilingual XNLI dataset\footnote{\href{https://huggingface.co/joeddav/xlm-roberta-large-xnli}{https://huggingface.co/joeddav/xlm-roberta-large-xnli}} to obtain NLI scores for pairs ``original news as premise $p$ $\leftrightarrow$ scraped news as hypothesis $h$''. Also, we generated input in a special format: 1) the premise was formulated as ``The news ``$<$news title + first $N$ symbols of content$>$" is legit"; 2) the hypothesis was only ``$<$news title + first $N$ symbols of content$>$". The size $N$ of the used content was a hyperparameter of this NLI-based approach for the news content similarity computation.

\subsubsection{Additional features}

\paragraph{Source credibility} As it was discussed in Section \ref{sec:fakenews_users}, one of the aspects to which users pay attention during news verification is the credibility of the news source. In addition, such a feature about external sources was widely used in methods described in Section \ref{sec:fakenews_review_methods}. We as well took into account the credibility of the source from which the piece of news comes. Following \cite{popat2016credibility}, we used AlexaRank for source assessment.

\paragraph{Named Entity frequency} During the manual experiment, it was discovered that cross-lingual check is more relevant for news about worldwide important events, people, or organizations and not the local ones. As a result, to evaluate the worthiness of the news to be cross-lingual checked we: 1) extracted NE from the title and the content of news; 2) found the most relevant page on Wikipedia; 3) evaluate AlexaRank of corresponding Wikipedia page to estimate the popularity of the NE.

\subsection{Comparison with Manual Markup}
To understand the validity of chosen approaches for content similarity computation between news, we conducted a small case study on a manually marked-up dataset. For each approach of news similarity estimation, we calculated the accuracy of such an experimental setup: the classification task if the scraped news supports the original news. So, from manually marked-up data we got a dataset of labeled 1000 pairs ``original news $\leftrightarrow$ scraped news". For each pair, we transferred from a three-person annotation to a single label by the voting of the majority.

Taking such a setup, we fined-tuned hyperparameters for both approaches. We fine-tuned threshold $\theta$ for the embeddings-based similarity. We conducted hyper-parameter search on the segment $[0.1, 0.9]$ with a step $\delta=0.1$. The best result was achieved with the $\theta=0.5$ threshold for decision making if the scraped news supports or not the original news. For NLI based approach, we fine-tuned the length of the text passed as the input to the NLI model. We got the best hyperparameters setup for the NLI approach is 500 symbols length of news text which is equal to the title of the news with the first two paragraphs of the content. For the NLI model, we united ``neutral" and ``contradiction" classes to have a similar setup as for the embeddings-based approach.

Finally, for \textit{cosine distance} approach we achieved $82\%$ accuracy, while for \textit{NLI} approach $70\%$ accuracy on 1000 pairs dataset. Although the models are not ideal, we believe that they can be used as baseline approximations of human judgments.

\subsection{Automatic Fake News Detection}
Finally, we conducted a set of experiments to validate Hypothesis \ref{hyp:third}: if the presented cross-lingual evidence feature can improve automatic fake news detection systems. We integrated the automated cross-lingual evidence feature into the fake news classification pipeline tested on three datasets.

\subsubsection{Datasets}
In tested datasets for our automated experiment, we tried to cover several worldwide spread topics -- politics, famous people and events, entertainment as well as the most recent event connected with COVID-19.. Firstly, we evaluate the systems on a multi-domain dataset by \cite{perez-rosas-etal-2018-automatic} which consist of two parts: \textit{FakeNewsAMT} dataset (240 fake and 240 legit articles) and \textit{CelebrityDataset} dataset (250 fake and 250 legit articles). \textit{FakeNewsAMT} dataset consists of news from six topics: sports, business, entertainment, politics, technology, and education. \textit{CelebrityDataset} is dedicated to rumors, hoaxes, and fake reports about famous actors, singers, socialites, and politicians.
Secondly, we ran experiments on COVID-19 fake news dataset \textit{ReCOVery} \cite{zhou2020recovery}. It consists of 2029 (665 fake and 1364 true news). All datasets are originally in English.
%The dataset was collected from sources of different level of reliability during January-May, 2020. 
%

\begin{table*}[h!]
    \centering
    \footnotesize
    \begin{tabular}{p{2.5cm}|c|c|p{10cm}}
        \toprule
        Dataset & \# Fakes & \# Legit & Covered topics \\
        \hline
        FakeNewsAMT & 240 & 240 & sports, business, entertainment, politics, technology, and education \\
        \hline
        CelebrityDataset & 250 & 250 & rumors, hoaxes, and fake reports about famous actors, singers, socialites, and politicians \\
        \hline
        ReCOVery & 665 & 1364 & rumors, hoaxes, and fake news about COVID-19 \\
        \bottomrule
    \end{tabular}
    \caption{Statics of datasets that were used to test fake news classification with proposed cross-lingual evidence feature.}
    \label{tab:fakenews_datasets_statistics}
\end{table*}

We used 70\%-20\%-10\% proportion for train-test-dev validation split.
 
\subsubsection{Baselines}
We compared our approach with several baselines. For the baseline, we chose the fake news systems based on internal features computed either via linguistic analysis or neural networks.

\textbf{Linguistic Features}: In \cite{perez-rosas-etal-2018-automatic} a baseline fake news classification model was trained based on Ngrams, punctuation, psycholinguistic features extracted with LIWC, readability, and syntax. In \cite{zhou2020recovery} LIWC features were also used as one of the proposed baselines. We tested these features separately, grouped them all, and in combination with our proposed feature. We experimented with SVM, RandomForest, LogRegression, and LightGBM. We used standard hyperparameters set for the models. The results of the best models based on LightGBM are presented. We call the model based on the concatenation of all listed above linguistic features as \textbf{All linguistic}.

\textbf{Text-CNN, LSTM}: Following \cite{zhou2020recovery}, we tested classical model for text categorization TextCNN and LSTM on all datasets.
%\cite{shu2019defend}

\textbf{BERT, RoBERTa}: BERT \cite{devlin-etal-2019-bert} based models were used for fake news detection by \cite{kaliyar2021fakebert} and specifically for COVID-19 fake news classification \cite{gundapu2021transformer,glazkova2020g2tmn}. We used pretrained models -- \textsf{bert-base-uncased}\footnote{\url{https://huggingface.co/bert-base-uncased}} and \textsf{roberta-base}\footnote{\url{https://huggingface.co/roberta-base}} -- and fine-tuned them.

\textbf{Only monolingual evidence (ME)}: In addition, we compared our feature with the case when only monolingual English evidence was used. For this baseline, the LightGBM model was used as well.

\subsubsection{Results}
To evaluate the performance of fake news classification models, we use three standard metrics: $F_1$, $precision$, $recall$. The formulas are provided bellow.
\begin{equation}
    precision = \frac{TP}{TP + FP}, ~~~ recall = \frac{TP}{TP + FN} ~~~
    F_1 = \frac{2 \cdot precision \cdot recall}{precision + recall}
\end{equation}

We experimented with both types of content similarity measurements -- either cosine similarity between embeddings (Emb.) or NLI scores -- concatenated with the source credibility rank (Rank) of the scraped news. Both Emb. and NLI features were presented as a vector of similarity scores for the pairs ``original news $\leftrightarrow$ scraped news''. 

\begin{figure}[h!]
\centering
\begin{tikzpicture}
  \centering
%   \hspace{-0.5cm}
  \begin{axis}[
        ybar, axis on top,
        height=8cm, width=21cm,
        bar width=0.3cm,
        ymajorgrids, tick align=inside,
        major grid style={draw=white},
        enlarge y limits={value=.1,upper},
        ymin=0, ymax=100,
        ytick={20,40,60,80,100},
        axis x line*=bottom,
        axis y line*=right,
        y axis line style={opacity=0},
        tickwidth=0pt,
        enlarge x limits=true,
        legend style={
            at={(0.5,-0.2)},
            anchor=north,
            legend columns=-1,
            /tikz/every even column/.append style={column sep=0.5cm}
        },
        symbolic x coords={
           BERT, RoBERTa, Ngrams, Punct., LIWC, Read., Syntax, All ling.},
       xtick=data,
    %   nodes near coords={
    %     \pgfmathprintnumber[precision=0]{\pgfplotspointmeta}
    %   },
       scale=0.75
    ]
    \addplot [draw=none, fill=blue!30] coordinates {
      (BERT, 58.6) (RoBERTa, 65.6) (Ngrams, 57.2) (Punct., 32.1) (LIWC, 59.2) (Read., 72.9) (Syntax, 62.4) (All ling., 73.9) };
   \addplot [draw=none,fill=red!30] coordinates {
      (BERT, 54.1) (RoBERTa, 87.2) (Ngrams, 65.5) (Punct., 74.1) (LIWC, 64.4) (Read., 76.0) (Syntax, 67.7) (All ling., 64.1) };
   \addplot [draw=none, fill=green!30] coordinates {
      (BERT, 89.4) (RoBERTa, 95.3) (Ngrams, 85.3) (Punct., 86.4) (LIWC, 88.4) (Read., 92.7) (Syntax, 89.5) (All ling., 93.7) };
      
    %\node[pin=45:{$e$}] at (axis cs:RoBERTa, 95.3) {F1 = 95.3};
    
    %\node[circle,fill=blue,scale=0.5,pin=135:{$(3,24)$}] at (axis cs:RoBERTa, 95.3) {F1 = 95.3};
      
    \node[above] at ($(axis cs:RoBERTa, 95.3)$) {\ \ \ \ \ \textcolor{red}{\textbf{95.3}}};
    \node[above] at ($(axis cs:All ling., 93.7)$) {\ \ \ \ \ \textbf{93.7}};

    \legend{Original, + CE AlexaRank, + CE Feature}
  \end{axis}
  \end{tikzpicture}
  \caption{Results on FakeNewsAMT dataset ($F_1$ score): adding proposed Cross-lingual Evidence (CE) improves various baseline systems and yields state-of-the-art results with RoBERTa model.}
  \label{fig:multilingual_fake_hist_fakenewsamt}
\end{figure}
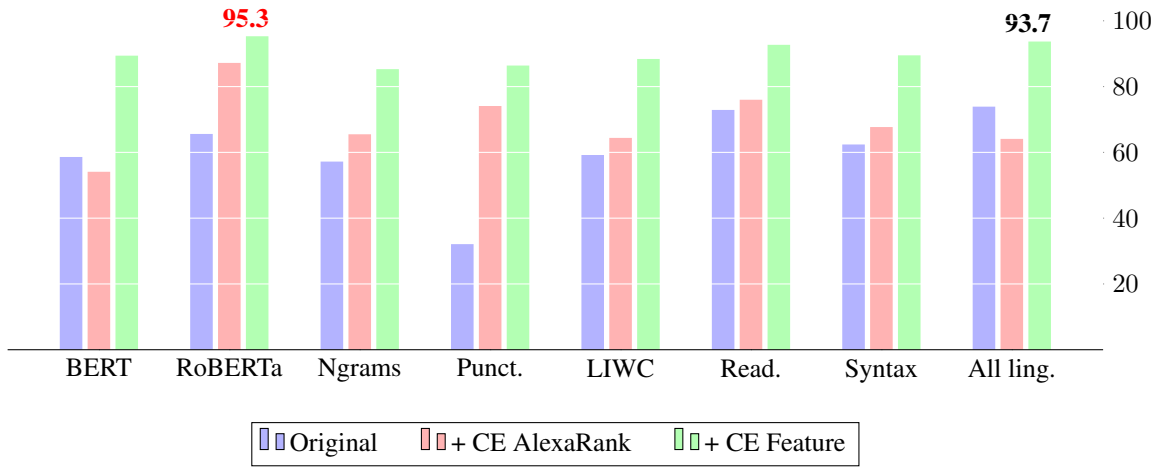

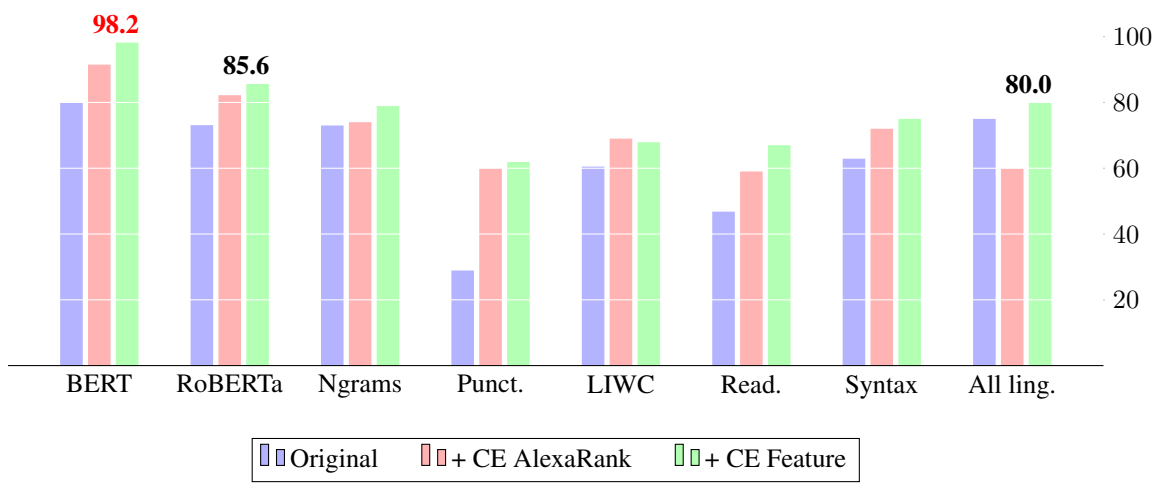
\begin{figure}[h]
\centering
\begin{tikzpicture}
  %\centering
  \hspace{-0.5cm}
  \begin{axis}[
        ybar, axis on top,
        height=8cm, width=21cm,
        bar width=0.3cm,
        ymajorgrids, tick align=inside,
        major grid style={draw=white},
        enlarge y limits={value=.1,upper},
        ymin=0, ymax=100,
        ytick={20,40,60,80,100},
        axis x line*=bottom,
        axis y line*=right,
        y axis line style={opacity=0},
        tickwidth=0pt,
        enlarge x limits=true,
        legend style={
            at={(0.5,-0.2)},
            anchor=north,
            legend columns=-1,
            /tikz/every even column/.append style={column sep=0.5cm}
        },
        symbolic x coords={
           BERT, RoBERTa, Ngrams, Punct., LIWC, Read., Syntax, All ling.},
       xtick=data,
    %   nodes near coords={
    %     \pgfmathprintnumber[precision=0]{\pgfplotspointmeta}
    %   },
       scale=0.75
    ]
    \addplot [draw=none, fill=blue!30] coordinates {
      (BERT, 80.0) (RoBERTa, 73.1) (Ngrams, 73.0) (Punct., 28.9) (LIWC, 60.5) (Read., 46.8) (Syntax, 62.9) (All ling., 75.0) };
   \addplot [draw=none,fill=red!30] coordinates {
      (BERT, 91.5) (RoBERTa, 82.2) (Ngrams, 74.0) (Punct., 60.0) (LIWC, 69.0) (Read., 59.0) (Syntax, 72.0) (All ling., 60.0) };
   \addplot [draw=none, fill=green!30] coordinates {
      (BERT, 98.2) (RoBERTa, 85.6) (Ngrams, 78.9) (Punct., 61.9) (LIWC, 67.9) (Read., 67.0) (Syntax, 75.0) (All ling., 80.0) };
      
    %\node[pin=45:{$e$}] at (axis cs:RoBERTa, 95.3) {F1 = 95.3};
    
    %\node[circle,fill=blue,scale=0.5,pin=135:{$(3,24)$}] at (axis cs:RoBERTa, 95.3) {F1 = 95.3};
      
    \node[above] at ($(axis cs:BERT, 98.2)$) {\ \ \ \ \ \textcolor{red}{\textbf{98.2}}};
    \node[above] at ($(axis cs:RoBERTa, 85.6)$) {\ \ \ \ \ \textbf{85.6}};
    \node[above] at ($(axis cs:All ling., 80.0)$) {\ \ \ \ \ \textbf{80.0}};

    \legend{Original, + CE AlexaRank, + CE Feature}
  \end{axis}
  \end{tikzpicture}
\caption{Results on Celebrity dataset ($F_1$ score): adding our Cross-lingual Evidence (CE) improves various baseline systems and yields state-of-the-art result with BERT model.}
\label{fig:multilingual_fake_hist_celebrity}
\end{figure}

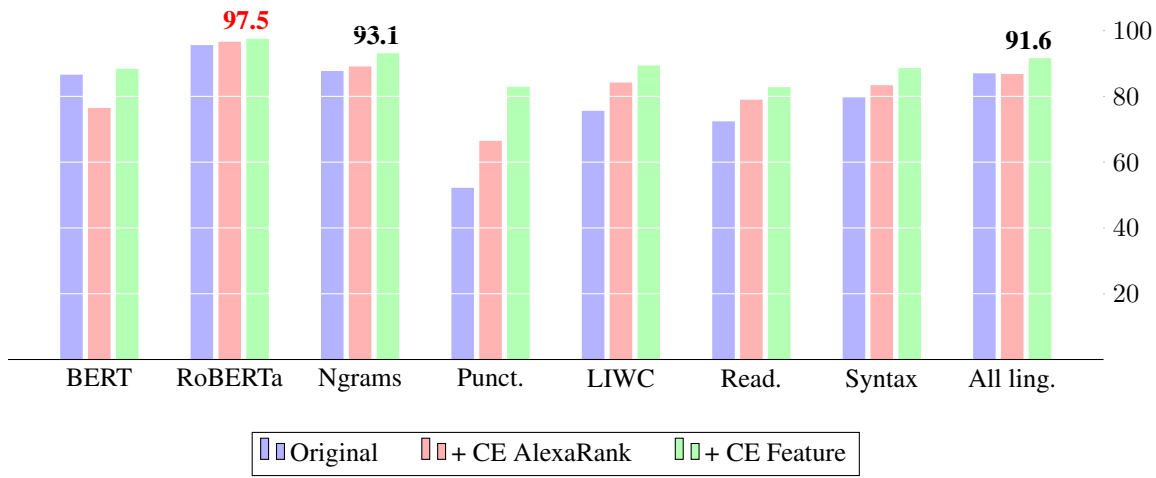
\begin{figure}[h!]
\centering
\begin{tikzpicture}
  %\centering
  \hspace{-0.5cm}
  \begin{axis}[
        ybar, axis on top,
        height=8cm, width=21cm,
        bar width=0.3cm,
        ymajorgrids, tick align=inside,
        major grid style={draw=white},
        enlarge y limits={value=.1,upper},
        ymin=0, ymax=100,
        ytick={20,40,60,80,100},
        axis x line*=bottom,
        axis y line*=right,
        y axis line style={opacity=0},
        tickwidth=0pt,
        enlarge x limits=true,
        legend style={
            at={(0.5,-0.2)},
            anchor=north,
            legend columns=-1,
            /tikz/every even column/.append style={column sep=0.5cm}
        },
        symbolic x coords={
           BERT, RoBERTa, Ngrams, Punct., LIWC, Read., Syntax, All ling.},
       xtick=data,
    %   nodes near coords={
    %     \pgfmathprintnumber[precision=0]{\pgfplotspointmeta}
    %   },
       scale=0.75
    ]
    \addplot [draw=none, fill=blue!30] coordinates {
      (BERT, 86.6) (RoBERTa, 95.6) (Ngrams, 87.7) (Punct., 52.2) (LIWC, 75.6) (Read., 72.4) (Syntax, 79.7) (All ling., 87.0) };
   \addplot [draw=none,fill=red!30] coordinates {
      (BERT, 76.5) (RoBERTa, 96.6) (Ngrams, 89.1) (Punct., 66.5) (LIWC, 84.2) (Read., 79.0) (Syntax, 83.4) (All ling., 86.8) };
   \addplot [draw=none, fill=green!30] coordinates {
      (BERT, 88.4) (RoBERTa, 97.5) (Ngrams, 93.1) (Punct., 82.9) (LIWC, 89.4) (Read., 82.8) (Syntax, 88.6) (All ling., 91.6) };
      
    %\node[pin=45:{$e$}] at (axis cs:RoBERTa, 95.3) {F1 = 95.3};
    
    %\node[circle,fill=blue,scale=0.5,pin=135:{$(3,24)$}] at (axis cs:RoBERTa, 95.3) {F1 = 95.3};
      
    \node[above] at ($(axis cs:RoBERTa, 97.5)$) {\ \ \ \ \ \textcolor{red}{\textbf{97.5}}};
    \node[above] at ($(axis cs:All ling., 91.6)$) {\ \ \ \ \ \textbf{91.6}};
    \node[above] at ($(axis cs:Ngrams, 93.1)$) {\ \ \ \ \ \textbf{93.1}};

    \legend{Original, + CE AlexaRank, + CE Feature}
  \end{axis}
  \end{tikzpicture}
\caption{Results on ReCOVert dataset ($F_1$ score): adding our Cross-lingual Evidence (CE) improves various baseline systems and yields state-of-the-art result with RoBERTa model.}
\label{fig:multilingual_fake_hist_recovery}
\end{figure}

Table \ref{tab:fakenews_results} compares the results of our model based on cross-lingual evidence (CE) with the baselines on three datasets. To prove the statistical significance of the result we used paired t-test on 5-fold cross-validation. All improvements presented in the results are statistically important. Additionally, we provide histogram view of $F_1$ scores comparison for all three datasets: FakeNewsAMT (Figure~\ref{fig:multilingual_fake_hist_fakenewsamt}), Celebrity (Figure~\ref{fig:multilingual_fake_hist_celebrity}), and ReCOVery (Figure~\ref{fig:multilingual_fake_hist_recovery}).

CE features along slightly outperform the baselines or show almost the same results as linguistic features. As it was expected, only ME based fake news detection system shows worse results than the usage of CE features. NLI based CE features show generally worse results than embeddings based approach. For further improvements, the NLI model can be additionally trained specifically for the task of detection of confirmation or refutation specifically in news content. 

The addition of the CE feature improves all baseline models. For \textit{FakeNewsAMT}, the best $F_1~=~0.973$ score is achieved with BERT embeddings in combination with CE features. For \textit{Celebrity} dataset, BERT again with CE features shows the best results achieving the best $F_1~=~982$ result. In spite RoBERTa showing the highest $F_1~=~0.975$ score for \textit{ReCOVery}, the combination of all linguistic and CE features and specifically Ngrams with CE features show competitive results achieving $F_1~=~0.916$ and $F_1~=~0.931$ respectively.

The importance of the proposed features in the model's decision-making is also confirmed by the feature's importance. The top-30 features' importance for best models for all datasets based on embeddings similarities is reported in Appendix~\ref{app:multilingual_fake_features}. For all \textit{FakeNewsAMT}, \textit{Celebrity}, and \textit{ReCOVery} dataset we can see the presence not only English, but indeed cross-lingual evidence features in the top important features. Although English evidence features for the top-3 news from the search results got the highest importance, the similarity scores and rank of the source from other languages (French, German, Spanish, Russian) contribute as well.

\begin{table}[ht!]
    \footnotesize
    \centering
    \begin{tabular}{p{4cm}p{0.8cm}p{0.8cm}p{0.8cm}p{0.5mm}p{0.8cm}p{0.8cm}p{0.8cm}p{0.5mm}p{0.8cm}p{0.8cm}p{0.8cm}}
    \toprule
         \phantom{abc} & \multicolumn{3}{c}{\textbf{FakeNewsAMT}} & \phantom{abc} & \multicolumn{3}{c}{\textbf{Celebrity}} & \phantom{abc} & \multicolumn{3}{c}{\textbf{ReCOVery}} \\
         \cmidrule{2-4} \cmidrule{6-8} \cmidrule{10-12}
         \phantom{abc} & Pre. & Rec. & F1 & \phantom{abc} & Pre. & Rec. & F1 & \phantom{abc} & Pre. & Rec. & F1 \\
         \midrule
         TextCNN & 0.276 & 0.250 & 0.260 & \phantom{abc} & 0.641 & 0.703 & 0.664 & \phantom{abc} & 0.733 & 0.913 & 0.805 \\
         LSTM & 0.614 & 0.614 & 0.614 & \phantom{abc} & 0.745 & 0.740 & 0.740 & \phantom{abc} & 0.800 & 0.803 & 0.793 \\
         ME Emb. + Rank & 0.539 & 0.593 & 0.592 & \phantom{abc} & 0.552 & 0.550 & 0.550 & \phantom{abc} & 0.794 & 0.798 & 0.793 \\
         ME NLI + Rank & 0.637 & 0.633 & 0.634 & \phantom{abc} & 0.554 & 0.550 & 0.550 & \phantom{abc} & 0.756 & 0.761 & 0.752 \\
         CE Emb. + Rank & 0.872 & 0.864 & 0.864 & \phantom{abc} & 0.631 & 0.620 & 0.619 & \phantom{abc} & 0.829 & 0.829 & 0.829 \\
         CE NLI + Rank & 0.837 & 0.833 & 0.834 & \phantom{abc} & 0.625 & 0.620 & 0.620 & \phantom{abc} & 0.767 & 0.771 & 0.762 \\
        \midrule
        
        BERT & 0.586 & 0.586 & 0.586 & \phantom{abc} & 0.800 & 0.800 & 0.800 & \phantom{abc} & 0.868 & 0.868 & 0.866 \\
        %  BERT + CE AlexaRank & 0.541 & 0.541 & 0.541 & \phantom{abc} & 0.810 & 0.728 & 0.915 & \phantom{abc} & 0.768 & 0.773 & 0.765 \\
         BERT + CE Emb + Rank & \textbf{0.884} & \textbf{0.885} & \textbf{0.894} & \phantom{abc} & \underline{\textbf{0.982}} & \underline{\textbf{0.982}} & \underline{\textbf{0.982}} & \phantom{abc} & \textbf{0.870} & \textbf{0.863} & \textbf{0.884} \\
         \midrule
         
         RoBERTa & 0.895 & 0.548 & 0.656 & \phantom{abc} & 0.856 & 0.690 & 0.731 & \phantom{abc} & 0.986 & 0.936 & 0.956 \\
        %  RoBERTa + CE AlexaRank & 0.930 & 0.820 & 0.872 & \phantom{abc} & 0.799 & \textbf{0.890} & 0.822 & \phantom{abc} & 0.949 & \underline{\textbf{0.986}} & 0.966 \\
         RoBERTa + CE Emb + Rank & \underline{\textbf{0.973}} & \underline{\textbf{0.938}} & \underline{\textbf{0.953}} & \phantom{abc} & \textbf{0.952} & 0.784 & \textbf{0.856} & \phantom{abc} & \underline{\textbf{0.992}} & 0.960 & \underline{\textbf{0.975}} \\
         \midrule
         
         Ngrams & 0.573 & 0.572 & 0.572 & \phantom{abc} & 0.730 & 0.730 & 0.730 & \phantom{abc} & {0.878} & {0.879} & {0.877} \\
        %  Ngrams + CE & &  &  & \phantom{abc} &  &  &  & \phantom{abc} &  &  &  \\
        %  \multicolumn{1}{l}{\hspace{0.5cm} Only rank} & 0.000 & 0.000 & 0.000 & \phantom{abc} & 0.000 & 0.000 & 0.000 & \phantom{abc} & 0.000 & 0.000 & 0.000 \\
         Ngrams + CE Emb. + Rank & \textbf{0.864} & \textbf{0.854} & \textbf{0.853} & \phantom{abc} & \textbf{0.789} & \textbf{0.790} & \textbf{0.789} & \phantom{abc} & \textbf{0.931} & \textbf{0.932} & \textbf{0.931} \\
         Ngrams + CE NLI + Rank & 0.844 & 0.844 & 0.844 & \phantom{abc} & 0.690 & 0.690 & 0.690 & \phantom{abc} & 0.862 & 0.860 & 0.856 \\
         \midrule
         
         Punctuation & 0.239 & 0.489 & 0.321 & \phantom{abc} & 0.211 & 0.460 & 0.289 & \phantom{abc} & 0.433 & 0.658 & 0.522 \\
        %  Punctuation + CE & &  &  & \phantom{abc} &  &  &  & \phantom{abc} &  &  &  \\
        %  \multicolumn{1}{l}{\hspace{0.5cm} Only rank} & 0.000 & 0.000 & 0.000 & \phantom{abc} & 0.000 & 0.000 & 0.000 & \phantom{abc} & 0.000 & 0.000 & 0.000 \\
          Punctuation + CE Emb. + Rank & \textbf{0.872} & 0.864 & 0.864 & \phantom{abc} & 0.631 & 0.620 & 0.619 & \phantom{abc} & \textbf{0.829} & \textbf{0.829} & \textbf{0.829} \\
          Punctuation + CE NLI + Rank & 0.870 & \textbf{0.865} & \textbf{0.865} & \phantom{abc} & \textbf{0.690} & \textbf{0.690} & \textbf{0.690} & \phantom{abc} & 0.767 & 0.771 & 0.762 \\
        \midrule
         
         LIWC & 0.597 & 0.593 & 0.592 & \phantom{abc} & 0.630 & 0.610 & 0.605 & \phantom{abc} & 0.768 & 0.771 & 0.756 \\
        %  LIWC + CE & &  &  & \phantom{abc} &  &  &  & \phantom{abc} &  &  &  \\
        %  \multicolumn{1}{l}{\hspace{0.5cm} Only rank} & 0.000 & 0.000 & 0.000 & \phantom{abc} & 0.000 & 0.000 & 0.000 & \phantom{abc} & 0.000 & 0.000 & 0.000 \\
         LIWC + CE Emb. + Rank & \textbf{0.894} & \textbf{0.885} & \textbf{0.884} & \phantom{abc} & \textbf{0.692} & \textbf{0.680} & \textbf{0.679} & \phantom{abc} & \textbf{0.894} & \textbf{0.894} & \textbf{0.894} \\ 
         LIWC + CE NLI + Rank & 0.850 & 0.844 & 0.844 & \phantom{abc} & 0.650 & 0.650 & 0.650 & \phantom{abc} & 0.816 & 0.815 & 0.808 \\
         \midrule 
         
         Readability & 0.729 & 0.729 & 0.729 & \phantom{abc}&  0.478 & 0.470 & 0.468 & \phantom{abc} & 0.732 & 0.741 & 0.724 \\
        %  Readability + CE & &  &  & \phantom{abc} &  &  &  & \phantom{abc} &  &  &  \\
        %  \multicolumn{1}{l}{\hspace{0.5cm} Only rank} & 0.000 & 0.000 & 0.000 & \phantom{abc} & 0.000 & 0.000 & 0.000 & \phantom{abc} & 0.000 & 0.000 & 0.000 \\
         Readability + CE Emb.+ Rank & \textbf{0.928} & \textbf{0.927} & \textbf{0.927} & \phantom{abc} & \textbf{0.674} & \textbf{0.670} & \textbf{0.670} & \phantom{abc} & \textbf{0.828} & \textbf{0.829} & \textbf{0.828} \\
         Readability + CE NLI + Rank & 0.854 & 0.854 & 0.854 & \phantom{abc} & 0.601 & 0.600 & 0.599 & \phantom{abc} & 0.772 & 0.773 & 0.762 \\
         \midrule 
         
         Syntax & 0.626 & 0.625 & 0.624 & \phantom{abc} & 0.639 & 0.630 & 0.629 & \phantom{abc} & 0.812 & 0.809 & 0.797 \\
        %  Syntax + CE & &  &  & \phantom{abc} &  &  &  & \phantom{abc} &  &  &  \\
        %  \multicolumn{1}{l}{\hspace{0.5cm} Only rank} & 0.000 & 0.000 & 0.000 & \phantom{abc} & 0.000 & 0.000 & 0.000 & \phantom{abc} & 0.000 & 0.000 & 0.000 \\
         Syntax + CE Emb. + Rank & \textbf{0.902} & \textbf{0.895} & \textbf{0.895} & \phantom{abc} & \textbf{0.754} & \textbf{0.750} & \textbf{0.750} & \phantom{abc} & \textbf{0.886} & \textbf{0.886} & \textbf{0.886} \\
         Syntax + CE  NLI + Rank & 0.505 & 0.500 & 0.501 & \phantom{abc} & 0.525 & 0.520 & 0.519 & \phantom{abc} & 0.840 & 0.837 & 0.832 \\
         \midrule 
         
         All linguistic & {0.739} & {0.739} & {0.739} & \phantom{abc} & {0.750} & {0.750} & {0.750} & \phantom{abc} & 0.875 & 0.874 & 0.870 \\
        %  All linguistic + CE & &  &  & \phantom{abc} &  &  &  & \phantom{abc} &  &  &  \\
        %  \multicolumn{1}{l}{\hspace{0.5cm} Only rank} & 0.000 & 0.000 & 0.000 & \phantom{abc} & 0.000 & 0.000 & 0.000 & \phantom{abc} & 0.000 & 0.000 & 0.000 \\
         All linguistic + CE Emb. + Rank & \underline{\textbf{0.940}} & \underline{\textbf{0.937}} & \underline{\textbf{0.937}} & \phantom{abc} & {\textbf{0.801}} & \underline{\textbf{0.800}} & \underline{\textbf{0.800}} & \phantom{abc} & \textbf{0.916} & \textbf{0.917} & \textbf{0.916} \\
         All linguistic + CE NLI + Rank & 0.886 & 0.885 & 0.886 & \phantom{abc} & 0.737 & 0.732 & 0.732 & \phantom{abc} & 0.864 & 0.865 & 0.862 \\
         %\midrule 
    \bottomrule
    \end{tabular}
    \caption{Results of integration of cross-lingual evidence (CE) feature into automated fake news classification systems. The proposed feature is used in two way based on content similarity computation strategy: (i) based on text embeddings (Emb.) (ii) based on NLI scores (NLI). It is also combined with the rank of the news articles source (Rank). The CE feature alongside showed worse results then baseline methods. All the improvements of the results were statistically proven by t-test on 5-fold cross-validation. However, in combination with linguistic features the SOTA results are achieved.} %TextCNN~\cite{textcnn} and another-strong-sota~\cite{bkljsdf}.}
    \label{tab:fakenews_results}
\end{table}

\subsection{Ablation study}
In order to verify, which part of the presented cross-lingual evidence feature impacted the results the most, we conducted ablation study.

We compared the best results of combinations of linguistic features and CE evidence features. We tested the usage of monolingual English evidence (ME) and only source ranks (Rank) in combination with linguistic features as well. The results are presented in Table \ref{tab:results_ablation}. 

We can see that the rank of cross-lingual evidence sources compared to the combination of content similarity and ranks performed worse. The same trend is observed for the combinations of linguistic features with ME Rank (rank of only English sources), ME Emb. + Rank (content similarity comparison based on embeddings between the original news and only English search results in combination of English sources' ranks), and CE Rank (the sources ranks of all scraped cross-lingual articles). For statistical significance proff for all comparisons with the best model we used paired t-test on 5-fold cross-validation as well. All the obtained results are statistically significant. Consequently, we can claim that the usage of proposed cross-lingual evidence feature is justified.

\begin{table*}[ht!]
    \footnotesize
    \centering
    \begin{tabular}{p{4cm}P{0.8cm}P{0.8cm}P{0.8cm}p{0.5mm}P{0.8cm}P{0.8cm}P{0.8cm}p{0.5mm}P{0.8cm}P{0.8cm}P{0.8cm}}
    \toprule
         \phantom{abc} & \multicolumn{3}{c}{\textbf{FakeNewsAMT}} & \phantom{abc} & \multicolumn{3}{c}{\textbf{Celebrity}} & \phantom{abc} & \multicolumn{3}{c}{\textbf{ReCOVery}} \\
         \cmidrule{2-4} \cmidrule{6-8} \cmidrule{10-12}
         \phantom{abc} & Pre. & Rec. & F1 & \phantom{abc} & Pre. & Rec. & F1 & \phantom{abc} & Pre. & Rec. & F1 \\
         
         \midrule
         CE Rank & 0.541 & 0.541 & 0.541 & \phantom{abc} & 0.605 & 0.605 & 0.605 & \phantom{abc} & 0.768 & 0.773 & 0.765 \\
         %CE NLI + Rank & 0.837 & 0.833 & 0.834 & \phantom{abc} & 0.625 & 0.620 & 0.620 & \phantom{abc} & 0.767 & 0.771 & 0.762 \\
         CE Emb. + Rank & \textbf{0.872} & \textbf{0.864} & \textbf{0.864} & \phantom{abc} & \textbf{0.631} & \textbf{0.620} & \textbf{0.619} & \phantom{abc} & \textbf{0.829} & \textbf{0.829} & \textbf{0.829} \\
        
        \midrule
         Ngrams + ME Rank & 0.646 & 0.645 & 0.644 & \phantom{abc} & 0.679 & 0.680 & 0.679 & \phantom{abc} & 0.802 & 0.802 & 0.800 \\
         Ngrams + ME Emb. + Rank & 0.656 & 0.656 & 0.656 & \phantom{abc} & 0.750 & 0.750 & 0.750 & \phantom{abc} & 0.808 & 0.807 & 0.805 \\ 
         %Ngrams + ME NLI + Rank & 0.844 & 0.844 & 0.844 & \phantom{abc} & 0.670 & 0.670 & 0.670 & \phantom{abc} & 0.866 & 0.865 & 0.861 \\ 
         Ngrams + CE Rank & 0.655 & 0.655 & 0.655 & \phantom{abc} & 0.740 & 0.740 & 0.740 & \phantom{abc} & 0.891 & 0.891 & 0.891 \\
         Ngrams + CE Emb. + Rank & \textbf{0.864} & \textbf{0.854} & \textbf{0.853} & \phantom{abc} & \textbf{0.789} & \textbf{0.790} & \textbf{0.789} & \phantom{abc} & \underline{\textbf{0.931}} & \underline{\textbf{0.932}} & \underline{\textbf{0.931}} \\
         
        \midrule
         Punct. + ME Rank & 0.604 & 0.604 & 0.603 & \phantom{abc} & 0.589 & 0.590 & 0.589 & \phantom{abc} & 0.718 & 0.721 & 0.717 \\
         %Punctuation + ME Emb. + Rank & 0.593 & 0.593 & 0.592 & \phantom{abc} & 0.000 & 0.000 & 0.000 & \phantom{abc} & 0.000 & 0.000 & 0.000 \\ 
         Punct. + ME NLI + Rank & 0.855 & 0.854 & 0.854 & \phantom{abc} & 0.670 & 0.670 & 0.670 & \phantom{abc} & 0.756 & 0.761 & 0.752 \\ 
         Punct. + CE Rank & 0.741 & 0.741 & 0.741 & \phantom{abc} & 0.605 & 0.600 & 0.600 & \phantom{abc} & 0.668 & 0.673 & 0.665 \\
         Punct. + CE NLI + Rank & \textbf{0.870} & \textbf{0.865} & \textbf{0.865} & \phantom{abc} & \textbf{0.690} & \textbf{0.690} & \textbf{0.690} & \phantom{abc} & 0.\textbf{767} & \textbf{0.771} & \textbf{0.762} \\ 
         
        \midrule
         LIWC + ME Rank & 0.646 & 0.645 & 0.643 & \phantom{abc} & 0.617 & 0.610 & 0.610 & \phantom{abc} & 0.771 & 0.771 & 0.769 \\
         LIWC + ME Emb. + Rank & 0.713 & 0.708 & 0.705 & \phantom{abc} & 0.643 & 0.640 & 0.640 & \phantom{abc} & 0.850 & 0.851 & 0.848 \\ 
         %LIWC + ME NLI + Rank & 0.849 & 0.844 & 0.844 & \phantom{abc} & 0.619 & 0.620 & 0.619 & \phantom{abc} & 0.823 & 0.823 & 0.816 \\ 
         LIWC + CE Rank & 0.646 & 0.645 & 0.644 & \phantom{abc} & 0.712 & 0.700 & 0.690 & \phantom{abc} & 0.846 & 0.846 & 0.842 \\
         LIWC + CE Emb. + Rank & \textbf{0.894} & \textbf{0.885} & \textbf{0.884} & \phantom{abc} & \textbf{0.692} & \textbf{0.680} & \textbf{0.679} & \phantom{abc} & \textbf{0.894} & \textbf{0.894} & \textbf{0.894} \\
         
        \midrule
         Read. + ME Rank & 0.650 & 0.650 & 0.650 & \phantom{abc} & 0.530 & 0.530 & 0.530 & \phantom{abc} & 0.797 & 0.801 & 0.796 \\
         Read. + ME Emb. + Rank & 0.739 & 0.739 & 0.739 & \phantom{abc} & 0.580 & 0.580 & 0.580 & \phantom{abc} & 0.808 & 0.811 & 0.806 \\ 
         %Readibility + ME NLI + Rank & 0.833 & 0.833 & 0.833 & \phantom{abc} & 0.647 & 0.640 & 0.639 & \phantom{abc} & 0.769 & 0.771 & 0.760 \\ 
         Read. + CE Rank & 0.760 & 0.760 & 0.760 & \phantom{abc} & 0.592 & 0.590 & 0.590 & \phantom{abc} & 0.796 & 0.798 & 0.790 \\
         Read. + CE Emb. + Rank & \textbf{0.928} & \textbf{0.927} & \textbf{0.927} & \phantom{abc} & \textbf{0.674} & \textbf{0.670} & \textbf{0.670} & \phantom{abc} & \textbf{0.828} & \textbf{0.829} & \textbf{0.828} \\
         
        \midrule
         Syntax + ME Rank & 0.670 & 0.666 & 0.663 & \phantom{abc} & 0.620 & 0.620 & 0.620 & \phantom{abc} & 0.754 & 0.754 & 0.749 \\
         Syntax + ME Emb. + Rank & 0.689 & 0.677 & 0.670 & \phantom{abc} & 0.656 & 0.650 & 0.650 & \phantom{abc} & 0.806 & 0.805 & 0.805 \\ 
         %Syntax + ME NLI + Rank & 0.510 & 0.510 & 0.510 & \phantom{abc} & 0.516 & 0.510 & 0.509 & \phantom{abc} & 0.842 & 0.840 & 0.834 \\ 
         Syntax + CE Rank & 0.677 & 0.677 & 0.677 & \phantom{abc} & 0.721 & 0.720 & 0.720 & \phantom{abc} & 0.844 & 0.841 & 0.834 \\
         Syntax + CE Emb. + Rank & \textbf{0.902} & \textbf{0.895} & \textbf{0.895} & \phantom{abc} & \textbf{0.754} & \textbf{0.750} & \textbf{0.750} & \phantom{abc} & \textbf{0.886} & \textbf{0.886} & \textbf{0.886} \\
         
        \midrule
         All ling. + ME Rank & 0.604 & 0.604 & 0.603 & \phantom{abc} & 0.589 & 0.590 & 0.589 & \phantom{abc} & 0.808 & 0.807 & 0.804 \\
         All ling. + ME Emb. + Rank & 0.803 & 0.802 & 0.801 & \phantom{abc} & 0.759 & 0.760 & 0.759 & \phantom{abc} & 0.808 & 0.807 & 0.804 \\ 
         %All linguistic + ME NLI + Rank & 0.875 & 0.875 & 0.875 & \phantom{abc} & 0.730 & 0.730 & 0.730 & \phantom{abc} & 0.867 & 0.867 & 0.864 \\ 
         All ling. + CE Rank & 0.641 & 0.641 & 0.641 & \phantom{abc} & 0.605 & 0.600 & 0.600 & \phantom{abc} & 0.868 & 0.868 & 0.868 \\
         All ling. + CE Emb. + Rank & \underline{\textbf{0.940}} & \underline{\textbf{0.937}} & \underline{\textbf{0.937}} & \phantom{abc} & {\textbf{0.801}} & \underline{\textbf{0.800}} & \underline{\textbf{0.800}} & \phantom{abc} & \textbf{0.916} & \textbf{0.917} & \textbf{0.916} \\
    \bottomrule
    \end{tabular}
    \caption{Results of ablation study: the usage of the best feature sets with cross-lingual evidence (CE) and source rank (Rank) compared with the usage of monolingual evidence (ME) and the source rank alongside. We can see that the performance of only source ranks or monolingual evidence is significantly worse than the usage of our proposed feature for all datasets.}
    \label{tab:results_ablation}
\end{table*}

\subsection{Usage and Explainability}
% We explored the time and memory that can be used by proposed best models based on cross-lingual evidence in the real applications during inference. The detailed statistics is presented in Table \ref{tab:time_and_memory}. In terms of time, the major part is spent on the feature extraction, while the inference step works quite quickly. Obviously, the transformer-based models occupies quite a lot of memory, while linguistic features-based model requires significantly less memory.

% \begin{table}[h!]
%     \centering
%     \begin{tabular}{p{4cm}|c|c|c}
%         \hline
%         Model & Time: Feature Extraction (s) & Time: Inference (s) & Memory Usage \\
%         \hline
%         BERT + CE Emb + Rank & 100 & 0.04 & 2.23 Gb\\
%         RoBERTa + CE Emb + Rank & 120 & 0.04 & 2.5 Gb \\
%         All ling. + CE Emb + Rank & 80 & 0.03 & 480 Mb \\
%         \hline
%     \end{tabular}
%     \caption{Time (per 128 samples) and memory usage statistics of the best proposed models calculated at GPU Tesla K80.}
%     \label{tab:time_and_memory}
% \end{table}

Our proposed feature can be easily used to explain the decision of the model and provide evidence from different sources of confirmation or refutation of the verifiable news. For instance, the full-text evidence can be generated on obtained news to give the user at least the source of critical attitude to the original news. As the first step of such explanation, we can report the scraped news, the rank of the source and the approximation of similarity to the original news. The examples of such report are illustrated in Appendix \ref{app:mutlilingual_fake_case}, Tables \ref{table:multilingual_fake_case_fake} and \ref{table:multilingual_fake_case_legit}.

As it can be seen, the hypothesis confirms on the real news examples. For fake news example ``Lottery winner arrested for dumping \$200,000 of manure on ex-boss’ lawn" we can see different scraped information. Some of the articles explicitly refute the news and name it fake: the information from Politifact site ``Viral post that lottery winner was arrested for dumping manure on former boss’ lawn reeks of falsity" and Spanish list of fake news ``Estas son las 50 noticias falsas que tuvieron mayor éxito en Facebook en 2018". However, some of the scraped news really copied the original title in different languages (``Un gagnant de loterie arrêté pour avoir déversé 200 000\$ de fumier sur la pelouse de son ex-patron", \foreignlanguage{russian}{ПОБЕДИТЕЛЬ ЛОТЕРЕИ АРЕСТОВАН ЗА ТО, ЧТО ПОТРАТИЛ \$200.000, ЧТОБЫ СВАЛИТЬ ГОРУ НАВОЗА НА ГАЗОН}). But, we can see from the source ranks that these articles come from unreliable sources and the user should think critically about the read information. The other titles either correlates in topic but give absolutely different information that does not support the original one (``Lotto-Gewinner holt Mega-Jackpot und lässt 291 Millionen Dollar sausen \& Lottery winner takes MegaJackpot and drops") or do not correlate in any way with the input article (``Histoire de Suresnes — Wikipedia \& History of Suresnes -- Wikipedia"). As a result, the user should be critical of the information read, since the number of confirmations is quite small and there are even claims that the news is fake.

On the contrary to the fake news, the legit news ``Bubonic plague outbreak in Mongolia" received a big amount of support from all target languages. We can see the information about bubonic plague is presented in the first results cross-lingual: ``Bubonic plague: Case found in China's Inner Mongolia", ``Epidémie : des cas de peste détectés en Chine et en Mongolie \& Epidemic", ``Mongolei: 15-Jähriger an Beulenpest gestorben - DER SPIEGEL", ``BROTE DE PESTE BUBÓNICA EN MONGOLIA", ``\foreignlanguage{russian}{В Монголии произошла вспышка бубонной чумы}". Most importantly, the similarity between cross-lingual news content reinforced with the fact that it came from quite reliable sources. Consequently, as we can see quite enough cross-lingual support from reliable sources to the original news, the probability to believe in this information is quite high.

\section{Discussion and Future Work}
The proposed cross-lingual evidence feature implemented using described pipeline in Section \ref{sec:experiment2} can have several limitations. 

The main limitation is the work with world-wide important and world-wide spread events. Unfortunately, our approach will not be applicable for very local news connected with local areas or personalities. However, even if the news is only about local famous person, in any case it can be covered in news sources in neighbour countries. As a result, cross-lingual check with neighbour-countries languages can also be useful.

Secondly, the usage of Google services for search and translation steps can bring the bias from personalized system. We tried to avoid personalization in search by using incognito mode during experiments to hide search history and location parameter. Nevertheless, Google search can use meta information and adjust the resulted feed. On the one hand, the usage of Google services motivated by user search experience. On the other hand, the reproduction of such experiments can be quite difficult. In our future work we plan to overcome such issue in the experiments by using already presaved snapshots of search in the Internet for exact period of time.

As we used automated translation to get the queries for cross-lingual search, there can be another side of this automated translation application -- some Internet editions can use automated translation to get the duplication of the news in the target language. Moreover, the method of machine translation are becoming more and more advanced every year. As a result, we can get the repetition of the news in search results over the different languages. However, we believe that our proposed pipeline can handle such cases as we incorporated in our feature the source rank of the news. We believe, that for now reliable editions still self-processed text material in their language. But in future work the addition of detection of machine generated texts can be considered.

Another part that can be added to the proposed cross-lingual feature is the cross-checking of the information not only with the original news but also between scraped multilingual evidence. That can add additional signals to the information verification process and reveal new details. Moreover, in this work we used linguistic features calculated only for the original news. However, such features can also be added to the all scraped news in different languages when the appropriate methods will be implemented.

In our presented experiments the original news were presented only in English. Moreover, the datasets and information noise in the Internet are generally exist in bigger amount for English than for any other language. In future work the research should be done to test the proposed feature for news originally presented in different languages other than English. Also, the amount of scraped evidence for language should be somehow normalized according to the overall amount of news that appears for the language.

\section{Conclusion}
% We presented an approach for fake news detection based on cross-lingual evidence (CE) which provides a different perspective on the event across languages verified in two experiments.
% %The results of the manual evaluation on the 20 news dataset and automated evaluation on the FakeNewsDataset support the idea that information from multilingual sources can be a strong feature for news verification. Our approach is simple and unsupervised so it can be readily applied to all main languages where machine translation and pre-trained language models available (but also it can be used as a signal in more complex supervised fake news detection setups). 
% %After that we integrated proposed technique as additional feature to the baseline model. 
% Fake news classification model with CE significantly improves performance over various baselines and compares favorably to SOTA. Besides, the CE is interpretable as a user can check in which and how many languages a piece of given news was found. % building confidence in the correctness of the prediction.
% We release publicly all codes and data.\footnote{\url{https://anonymous.4open.science/r/5bad7b14-64bc-42ea-9549-da6c3074c6bc/}}
We presented \textsf{Multiverse} -- an approach for fake news detection based on cross-lingual evidence (CE) which is motivated by user behaviour and overcome the limitations of monolingual external featured of previous work. 

Firstly, we conducted manual study on 20 news datasets to test the hypothesis whether the real-life user can use cross-lingual evidence to detect fake news. The annotators successfully passed the task of such news verification providing also the markup of 100 pairs ``original news $\leftrightarrow$ scraped news". After the first hypothesis confirmation, we tested our approach for automated detection of fake news.

We experimented with two strategies for content similarity estimation: (i)~based on cosine distance between news texts embeddings; (ii)~based on natural language inference (NLI) scores where the original news used as premise $p$ and the scraped news as hypothesis $h$. We compared the proposed strategies with human assessments of 1000 pairs of marked news showing that these methods can be used for news similarity estimation. Finally, we integrated the proposed cross-lingual feature into automated fake news detection pipeline. To this moment, the cross-lingual feature itself showed the performance only at the baselines level. However, in combination with linguistic features based on the text original news it outperformed both statistical and deep learning fake news classification systems. 

Additionally, we provided ablation study where the necessity of the usage of cross-lingual evidence with source rank in comparison to only monolingual features was proven. In the end, we showed how the obtained cross-lingual information can be used for further evidence generation for the end users. 

\section{Acknowledgments*}
We would like to acknowledge the help of volunteers who participated in the user study presented in the first experiment in the article. 

%Bibliography
\bibliographystyle{unsrt}  
\bibliography{references}  

\clearpage

\appendix
\section{Feature Importance for Fake New Classification method}
\sectionmark{Feature Importance}
\label{app:multilingual_fake_features}

\begin{figure}[h!]
    \centering
    \includegraphics[width=\textwidth]{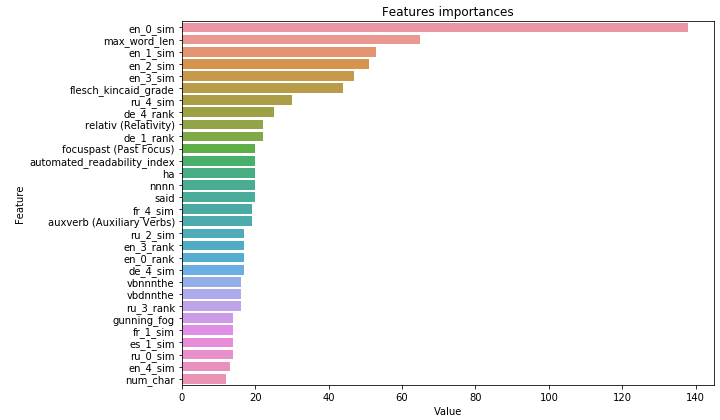}
    \caption{Top 30 features importances of the best model for FakeNewsAMT dataset: LightGBM model based on All linguistic + CE Emb. + Rank feature set.}
    \label{fig:multilingul_fake_fakenewsamt_features}
\end{figure}

In this section, we provide the illustration of feature importance for fake news classification model for: i)~FakeNewsAMT dataset (Figure~\ref{fig:multilingul_fake_fakenewsamt_features}); ii)~Celebrity dataset (Figure~\ref{fig:multilingul_fake_celebrity_features}); iii)~ReCOVery dataset (Figure~\ref{fig:multilingul_fake_celebrity_features}). The notation for CE feature designation: $<$language of news$>$\_$<$its position in search results$>$\_$<$content similarity feature (sim)$>$ or $<$source rank feature (rank)$>$. We can see that cross-lingual evidence features (both similarities and ranks) are at the top for all datasets.

\begin{figure}[h!]
    \centering
    \includegraphics[width=\textwidth]{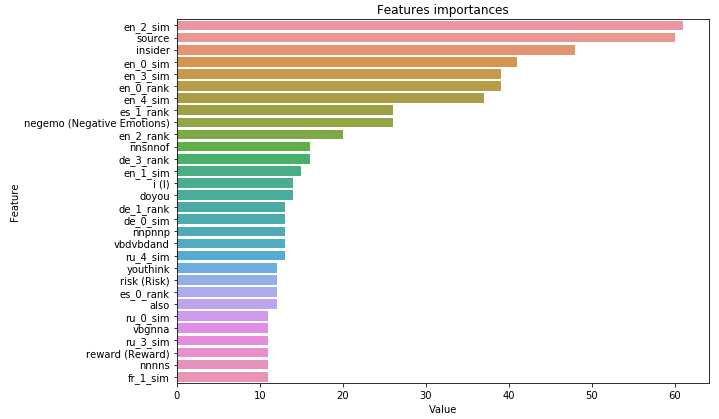}
    \caption{Top 30 features importances of the best model for Celebrity dataset: LightGBM model based on All linguistic + CE Emb. + Rank feature set.}
    \label{fig:multilingul_fake_celebrity_features}
\end{figure}

\begin{figure}[h!]
    \centering
    \includegraphics[width=\textwidth]{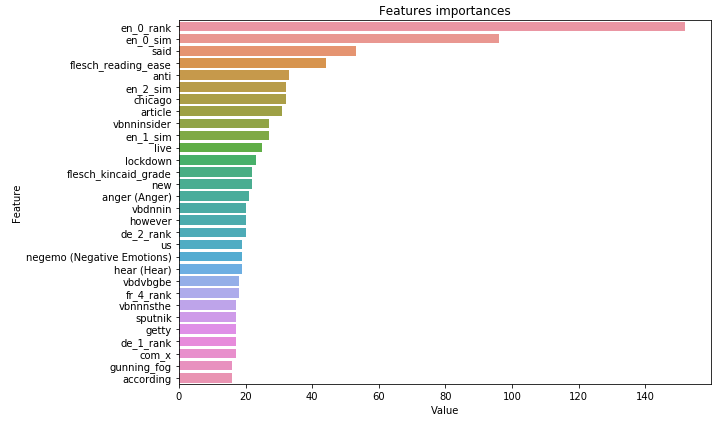}
    \caption{Top 30 features importances of the best model for ReCOVery dataset: LightGBM model based on Ngrams + CE Emb. + Rank feature set.}
    \label{fig:multilingul_fake_celebrity_features}
\end{figure}

\clearpage

\section{\textsf{Mutliverse} usage: Real-Case Example}
\label{app:mutlilingual_fake_case}
Here we provide examples of how the proposed \textsf{Mutliverse} approach for cross-lingual evidence feature extraction can be used for the explanation of fake news classification model decision explanation. In Table~\ref{table:multilingual_fake_case_fake}, we provide an example which cross-lingual evidence is extracted for \textbf{fake} news. We can observe that there is no supportive information and even refutation. On the contrary, for \textbf{legit} news we can observe a lot of supportive information all across different media.

\begin{table}[h!]
\footnotesize
\scriptsize
\begin{tabular}{p{6cm}|p{6cm}|p{2.5cm}|p{2.5cm}}
\toprule
\multicolumn{1}{c|}{\textbf{Title}} & \multicolumn{1}{c|}{\textbf{English translation}} & \multicolumn{1}{c|}{\textbf{\shortstack{Source \\ rank}$\downarrow$}} & \multicolumn{1}{c}{\textbf{\shortstack{Similarity \\ score}$\uparrow$}} \\
\hline
\multicolumn{4}{c}{\textbf{Original news (\color{red}{FAKE})}} \\
\hline
Lottery winner arrested for dumping \$200,000 of manure on ex-boss’ lawn & \multicolumn{1}{c|}{--} & \multicolumn{1}{c|}{--} & \multicolumn{1}{c}{--} \\
\hline

\multicolumn{4}{c}{\textbf{English search results}} \\
\hline
\rowcolor{Gray} PolitiFact - Viral post that lottery winner was arrested for dumping manure on former boss’ lawn reeks of falsity & \multicolumn{1}{c|}{--} & \multicolumn{1}{c|}{15947} & \multicolumn{1}{c}{0.00}\\
Was a Lottery Winner Arrested for Dumping \$200,000 of Manure on the Lawn of His Former Boss? & \multicolumn{1}{c|}{--} & \multicolumn{1}{c|}{5798} & \multicolumn{1}{c}{0.00}\\
\rowcolor{Gray}Lottery winner arrested for dumping \$200,000 of manure on ex-boss' lawn & \multicolumn{1}{c|}{--} & \multicolumn{1}{c|}{314849} & \multicolumn{1}{c}{0.89}\\
\hline

\multicolumn{4}{c}{\textbf{French search results}} \\
\hline
Un gagnant de loterie arrêté pour avoir déversé 200 000\$ de fumier sur la pelouse de son ex-patron | Africa24.info & Lottery winner arrested for dumping \$ 200,000 in manure on expatron's lawn Africa24info & \multicolumn{1}{c|}{2595725} & \multicolumn{1}{c}{0.78}\\
Fertiliser le jardin & Fertilize the garden & \multicolumn{1}{c|}{193218} & \multicolumn{1}{c}{0.43}\\
\rowcolor{Gray}Histoire de Suresnes — Wikipedia & History of Suresnes -- Wikipedia & \multicolumn{1}{c|}{13} & \multicolumn{1}{c}{0.31}\\
\hline

\multicolumn{4}{c}{\textbf{German search results}} \\
\hline
\rowcolor{Gray} Mit ``Scream"-Maske zum Millionen-Jackpot: Lottogewinner will anonym bleiben - aber er übersieht eine wichtige Sache & With a Scream mask for the millionaire jackpot lottery winner, he wants to remain anonymous but he overlooks an important thing & \multicolumn{1}{c|}{15294} & \multicolumn{1}{c}{0.55}\\
Lotto-Gewinner holt Mega-Jackpot und lässt 291 Millionen Dollar sausen & Lottery winner takes MegaJackpot and drops \$ 291 million & \multicolumn{1}{c|}{15294} & \multicolumn{1}{c}{0.58}\\
\rowcolor{Gray} Hesse knackt Sechs-Millionen-Jackpot: Noch hat sich der Gewinner nicht gemeldet & Hesse cracks six million jackpot The winner has not yet announced & \multicolumn{1}{c|}{44799} & \multicolumn{1}{c}{0.57}\\
\hline

\multicolumn{4}{c}{\textbf{Spanish search results}} \\
\hline
\rowcolor{Gray} Ganador de 125 millones en la lotería arrestado por vaciar camiones de heces en casa de su jefe & 125 million lottery winner arrested for dumping trucks of feces at his boss's home & \multicolumn{1}{c|}{922337} & \multicolumn{1}{c}{0.76}\\
Le toca la lotería y compra 20.000 toneladas de estiércol para arrojar en el porche de su jefe & He wins the lottery and buys 20,000 tons of manure to dump on his boss's porch & \multicolumn{1}{c|}{149185} & \multicolumn{1}{c}{0.77}\\
\rowcolor{Gray} Estas son las 50 noticias falsas que tuvieron mayor éxito en Facebook en 2018 & These are the 50 fake news that had the most success on Facebook in 2018 & \multicolumn{1}{c|}{405} & \multicolumn{1}{c}{0.00}\\
\hline

\multicolumn{4}{c}{\textbf{Russian search results}} \\
\hline
\rowcolor{Gray} \foreignlanguage{russian}{ПОБЕДИТЕЛЬ ЛОТЕРЕИ АРЕСТОВАН ЗА ТО, ЧТО ПОТРАТИЛ \$200.000, ЧТОБЫ СВАЛИТЬ ГОРУ НАВОЗА НА ГАЗОН / победитель :: смешные картинки (фото приколы) :: новости} & LOTTERY WINNER ARRESTED FOR SPENDING \$ 200,000 TO DUMP MOUNT OF MANURE ON THE LAWN / winner :: funny pictures (funny photos) :: news  & \multicolumn{1}{c|}{15418} & \multicolumn{1}{c}{0.76}\\
\foreignlanguage{russian}{ПОБЕДИТЕЛЬ ЛОТЕРЕИ АРЕСТОВАН ЗА ТО, ЧТО ПОТРАТИЛ \$200.000, ЧТОБЫ СВАЛИТЬ ГОРУ НАВОЗА НА ГАЗОН СВОЕГО БЫВШЕГО БОССА ПО НЕМУ ВИДНО, ЧТО ОНО ТОГО СТОИЛО...} & LOTTERY WINNER ARRESTED FOR SPENDING \$ 200,000 TO DUMP A MOUNTAIN OF MANURE ON THE LAW OF HIS FORMER BOSS ONE SEE THAT IT WAS WORTH ...  & \multicolumn{1}{c|}{146662} & \multicolumn{1}{c}{0.70}\\
\rowcolor{Gray} \foreignlanguage{russian}{Победитель лотереи потратил выигрыш, убойно отомстив бывшему боссу} & Lottery Winner Wasted Winning In Hellful Revenge On Ex-Boss & \multicolumn{1}{c|}{146662} & \multicolumn{1}{c}{0.83}\\
\bottomrule
\end{tabular}
\caption{The example of work of the proposed approach for fake and legit news. For each target language (English, French, German, Spanish, Russian) search results are presented: titles of top 3 news. For every non-Enlgish title the English translation is provided. Each piece of scraped news is rated with the rank of its source and content similarity to the original news based on text embedding. The larger$\uparrow$ (or lower$\downarrow$) score, the better. For \textbf{fake news} the search results either come from unreliable sources or provide no relevant information to the original news.}
\label{table:multilingual_fake_case_fake}
\end{table}

\begin{table}[h!]
\centering
\footnotesize
\begin{tabular}{p{6cm}|p{6cm}|p{2.5cm}|p{2.5cm}}
\toprule
\multicolumn{1}{c|}{\textbf{Title}} & \multicolumn{1}{c|}{\textbf{English translation}} & \multicolumn{1}{c|}{\textbf{\shortstack{Source \\ rank}$\downarrow$}} & \multicolumn{1}{c}{\textbf{\shortstack{Similarity \\ score}$\uparrow$}} \\
\hline
\multicolumn{4}{c}{\textbf{Original news (\color{blue}{LEGIT})}} \\
\hline
\foreignlanguage{russian}{В Монголии произошла вспышка бубонной чумы: \href{https://hightech.fm/2020/07/02/plague-outbreak}{https://hightech.fm/2020/07/02/plague-outbreak}} & Bubonic plague outbreak in Mongolia \\
\hline

\multicolumn{4}{c}{\textbf{English search results}} \\
\hline
\rowcolor{Gray} Bubonic plague: Case found in China's Inner Mongolia - CNN & \multicolumn{1}{c|}{--} & \multicolumn{1}{c|}{91} & \multicolumn{1}{c}{0.88}\\
Teenager dies of Black Death in Mongolia & \multicolumn{1}{c|}{--} & \multicolumn{1}{c|}{178} & \multicolumn{1}{c}{0.72}\\
\rowcolor{Gray} China bubonic plague: Inner Mongolia takes precautions after case & \multicolumn{1}{c|}{--} & \multicolumn{1}{c|}{101} & \multicolumn{1}{c}{0.69}\\
\hline

\multicolumn{4}{c}{\textbf{French search results}} \\
\hline
\rowcolor{Gray} Epidémie : des cas de peste détectés en Chine et en Mongolie & Epidemic: cases of plague detected in China and Mongolia & \multicolumn{1}{c|}{284} & \multicolumn{1}{c}{0.73}\\
Craintes d’une épidémie de peste bubonique? Un adolescent de 15 ans est la première victime recensée en Mongolie & Fear of a bubonic plague epidemic? A 15-year-old is the first victim in Mongolia & \multicolumn{1}{c|}{496} & \multicolumn{1}{c}{0.70}\\
\rowcolor{Gray} Chine : Un cas de peste bubonique détecté en Mongolie intérieure & China: Bubonic plague case detected in Inner Mongolia & \multicolumn{1}{c|}{5003} & \multicolumn{1}{c}{0.84}\\
\hline

\multicolumn{4}{c}{\textbf{German search results}} \\
\hline
\rowcolor{Gray} Mongolei: 15-Jähriger an Beulenpest gestorben - DER SPIEGEL & Mongolia: 15-year-old died of bubonic plague - DER SPIEGEL & \multicolumn{1}{c|}{928} & \multicolumn{1}{c}{0.78}\\
Beulenpest - Was über die Pest-Fälle in China bekannt & Bubonic plague - what is known about the plague cases in China & \multicolumn{1}{c|}{6234} & \multicolumn{1}{c}{0.75}\\
\rowcolor{Gray} Bringen Murmeltiere die Pest zurück? Mongolei warnt vor Tier-Kontakt & Will marmots bring the plague back? Mongolia warns of animal contact & \multicolumn{1}{c|}{48864} & \multicolumn{1}{c}{0.61}\\
\hline

\multicolumn{4}{c}{\textbf{Spanish search results}} \\
\hline
\rowcolor{Gray} BROTE DE PESTE BUBÓNICA EN MONGOLIA & BUBONIC PLAGUE OUTBREAK IN MONGOLIA & \multicolumn{1}{c|}{436} & \multicolumn{1}{c}{0.84}\\
Brote de peste negra provoca cuarentena en Mongola & Black plague outbreak causes quarantine in Mongolia & \multicolumn{1}{c|}{4417} & \multicolumn{1}{c}{0.78}\\
\rowcolor{Gray} Brote de peste negra alarma en Mongolia y cierra frontera con Rusia & Black plague outbreak alarms Mongolia, closes border with Russia & \multicolumn{1}{c|}{453} & \multicolumn{1}{c}{0.63}\\
\hline

\multicolumn{4}{c}{\textbf{Russian search results}} \\
\hline
\rowcolor{Gray} \foreignlanguage{russian}{В Монголии произошла вспышка бубонной чумы ... - Гордон} & There was an outbreak of bubonic plague in Mongolia ... - Gordon & \multicolumn{1}{c|}{21372} & \multicolumn{1}{c}{0.91}\\
\foreignlanguage{russian}{В Монголии произошла вспышка бубонной чумы - Урал56.Ру} & Bubonic plague outbreak in Mongolia - Ural56.Ru & \multicolumn{1}{c|}{124712} & \multicolumn{1}{c}{0.92}\\
\rowcolor{Gray} \foreignlanguage{russian}{Возвращение «Черной смерти»: главное о вспышке бубонной чумы в Монголии} & Return of the "Black Death": the main thing about the outbreak of the bubonic plague in Mongolia & \multicolumn{1}{c|}{8425} & \multicolumn{1}{c}{0.87}\\
\bottomrule
\end{tabular}
\caption{The example of work of the proposed approach for fake and legit news. For each target language (English, French, German, Spanish, Russian) search results are presented: titles of top 3 news. For every non-Enlgish title the English translation is provided. Each piece of scraped news is rated with the rank of its source and content similarity to the original news based on text embedding. The larger$\uparrow$ (or lower$\downarrow$) score, the better. For \textbf{legit news} the search results across different languages are strongly related to the original news.}
\label{table:multilingual_fake_case_legit}
\end{table}

\end{document}